\documentclass[final]{cvpr}

\usepackage{times}
\usepackage{epsfig}
\usepackage{graphicx}
\usepackage{amsmath}
\usepackage{amssymb}
\usepackage{listings}

\usepackage{algorithm}
\usepackage[noend]{algpseudocode}
\usepackage[caption=false, font=footnotesize, labelformat=empty]{subfig}
\usepackage{afterpage}
\usepackage{xcolor}
\usepackage{lscape}
\usepackage{fancyhdr}

\makeatletter
\def\BState{\State\hskip-\ALG@thistlm}
\makeatother

\usepackage[pagebackref=true,breaklinks=true,colorlinks,bookmarks=false]{hyperref}

\fancypagestyle{firstpage}
{
    \fancyfoot[L]{}    
    \fancyfoot[R]{\tiny This work has been submitted to the IEEE for possible publication. \\ Copyright may be transferred without notice, after which this version may no longer be accessible.}
    
}

\begin{document}

\title{Overparametrization of HyperNetworks at Fixed FLOP-Count \\ Enables Fast Neural Image Enhancement}

\author{Lorenz K. Muller\\
Huawei Technologies\\
Zurich Research Center, Switzerland\\
{\tt\small lorenz.mueller@huawei.com}}

\maketitle

\begin{abstract}
	Deep convolutional neural networks can enhance images taken with 
	small mobile camera sensors and excel at tasks like 
	demoisaicing, denoising and super-resolution. However, for practical 
	use on mobile devices these networks often require too many FLOPs and
	reducing the FLOPs of a convolution layer, also reduces its parameter count.
	This is problematic in view of the recent finding that heavily 
	over-parameterized neural networks are often the ones that generalize
	best. 
	
	In this paper we propose to use HyperNetworks to 
	break the fixed ratio of FLOPs to parameters of standard convolutions.
 	This allows us to 
	exceed previous state-of-the-art architectures in SSIM and MS-SSIM on the Zurich RAW-to-DSLR (ZRR) data-set at
	$>10 \times$ reduced FLOP-count. 
	On ZRR we further observe generalization curves consistent with `double-descent' behavior 
	at fixed FLOP-count, in the large image limit.	
	Finally we demonstrate
	the same technique can be applied to an existing network (VDN)
	to reduce its computational cost while maintaining fidelity
	on the Smartphone Image Denoising Dataset (SIDD).	
	
	Code for key functions is given in the appendix.

\end{abstract}
\thispagestyle{firstpage}
\section{Introduction}
\begin{figure}[ht]
\begin{center}
   \includegraphics[width=0.8\linewidth]{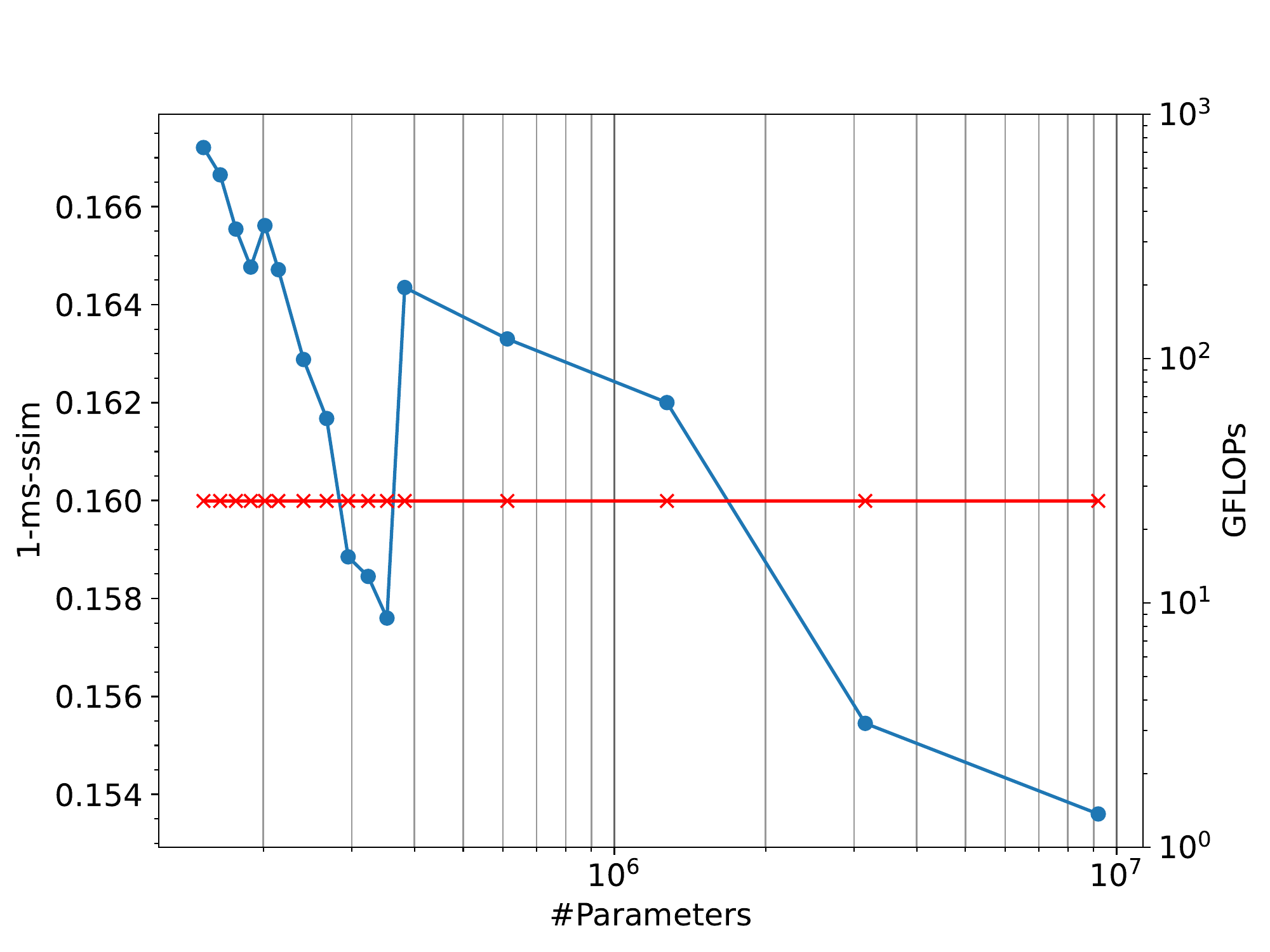}
   \end{center}
\begin{center}
	\includegraphics[width=0.3\linewidth]{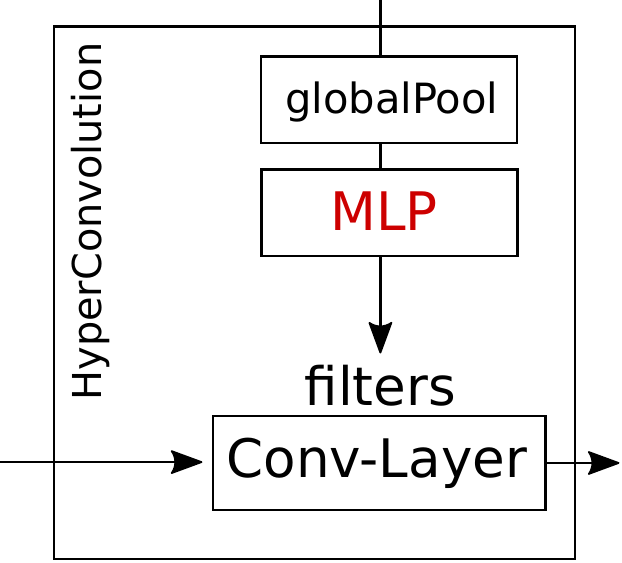}
	\includegraphics[width=0.5\linewidth]{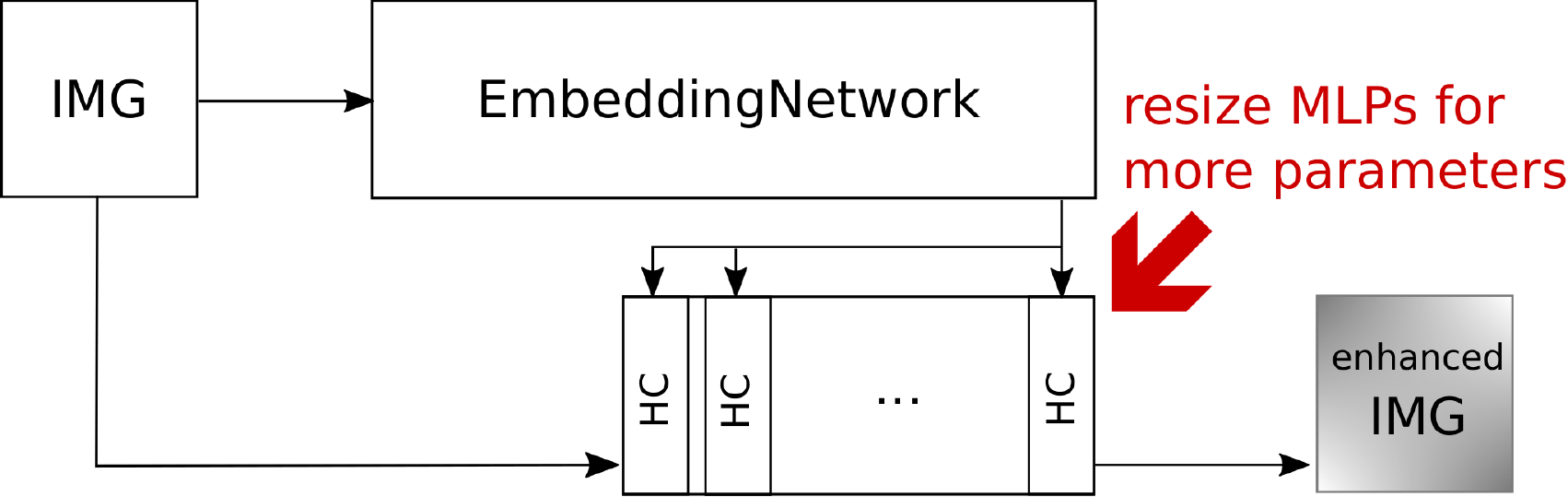}
\end{center}

   \caption{It is possible to improve generalization 
   in a ConvNet at near constant FLOP count with increasing over-parametrization, by
   using HyperConvolutions (HC) with predicted filters.
   Usually the ratio of parameters to FLOPs is fixed in 
   convolution layers. The MLPs in the HC can increase this ratio. }
\label{fig:concept}
\end{figure}
In recent years we have seen exceptional progress in computational image enhancement,
fueled among others by better processing pipelines \cite{brooks2019unprocessing}, problem formulations \cite{hu2018exposure}
and increasingly powerful
deep learning models \cite{ignatov2020replacing}. The resulting algorithms 
meet wide-spread interest in the era of mobile phone cameras,
whose sensors are small and therefore prone to producing
low-quality RAW data (compared to larger sensors).

For the mobile setting however, permissible power consumption is limited.
FLOP-count and memory use need to be kept on a budget for 
practical utility \cite{ignatov2018ai}. Similarly in the data-center setting,
power consumption and FLOP count of models is a growing concern,
for both economical as well as environmental reasons \cite{strubell2019energy}.
Unfortunately, in deep learning often larger models are better models \cite{nakkiran2019deep}.

This dilemma can be better understood given the context of two observations.
Firstly, in convolution layers FLOP count and parameter count are tied together.
The ratio between them is proportional to the input feature-map resolution (see also 
Sec.~\ref{sec:method}). Secondly, the classical bias-variance trade off \cite{bishop2006pattern}
does not fully characterize the generalization behavior of neural 
networks. For neural networks often optimal generalization 
occurs in the many parameter limit; i.e. one finds a `double-descent' generalization
curve \cite{belkin2018reconciling} on the plot of test-error vs. parameter count. 
 
If indeed computer vision tasks, and the 
neural networks used to tackle them, do exhibit this kind of generalization
behavior (as e.g. shown in \cite{nakkiran2019deep}), this spells a problem for constructing ConvNet
variants with few FLOPs, because it may be impossible to endow them
with sufficiently many parameters.

Fixing this problem by 
resizing images or giving up translation equivariance 
seems unappealing, because resolution reduction requires discarding 
information and translation equivariance is intuitively 
useful for many computer vision tasks (and necessary for some 
physics models).  

In this paper we propose and investigate an alternative approach 
to break the FLOP count to parameter count ratio, that preserves 
translation equivariance and image scale. This approach is
predicting convolutional filters from the input image, by
way of a HyperNetwork. We show that this allows for double-descent
generalization at fixed FLOP count (in the large input image limit)
on the example of a fully neural ISP for the Zurich RAW to DSLR data-set \cite{ignatov2020replacing}.

\section{Main Idea and Contributions}
\subsection{Main Idea}
The ratio of number of parameters to number of operations in a convolution layer 
is fixed at a given resolution. It is $2 \cdot H \cdot W$, where $H,W$ are the input height and width (see also Sec. \ref{sec:method}).

To break this dependency we can either re-scale the input or relax the weight-sharing
scheme of the convolution (not apply the same filter everywhere on the input).
Both of these approaches have a significant impact on the inductive bias of the 
resulting layer: Re-scaling the input decreases spatial resolution and relaxing
weight-sharing disrupts translation-equivariance. 

In this paper we propose a third avenue for modifying this fixed ratio. We use 
meta-parametrization, parameters that are themselves the results of computations.
A hyper-network \cite{ha2016hypernetworks} predicts the convolutional filters of the forward network 
for each input image. This hyper-network can contain global pooling and fully-connected layers.
By re-scaling these fully-connected layers, we can change the number of parameters of the overall 
network. Because the fully-connected layers do not scale with input image size, in 
the limit of large images, their impact on FLOP count is negligible. See Fig.
\ref{fig:concept} for an illustration of the main idea.

As a result we can build networks that simultaneously  1) operate at full resolution,
2) keep translation equivariance,
3) have high parameter count,
4) have low FLOP count.

\subsection{Contributions}
The main contributions of this paper can be summarized as follows:
\begin{itemize}
\item We identify the problem of fixed FLOP to parameter ratio for building 
efficient ConvNets.
\item We investigate generalization behavior as a function of parameter count at fixed\footnote{in the large 
image limit} FLOP count in ConvNets (Sec. \ref{subsec:dd}).
\item We outperform previous state-of-the-art architectures in SSIM \cite{dai2020awnet} and MS-SSIM \cite{ignatov2020replacing} on the Zurich RAW to DSLR task 
at $>10\times$ fewer FLOPs (Sec. \ref{subsec:large})
\item We reduce the FLOP count of the well-established VDN \cite{yue2019variational} by $>6\times$ without loss in fidelity on the SIDD benchmark \cite{abdelhamed2018high} (Sec. \ref{subsec:SIDD}) with a `drop-in' replacement for convolution layers (code in appendix).
\end{itemize}

\section{Related Work}

\subsection{Double Descent Generalization}
\label{sec:dd}
\begin{figure}[th]
\begin{center}
   \includegraphics[width=0.8\linewidth]{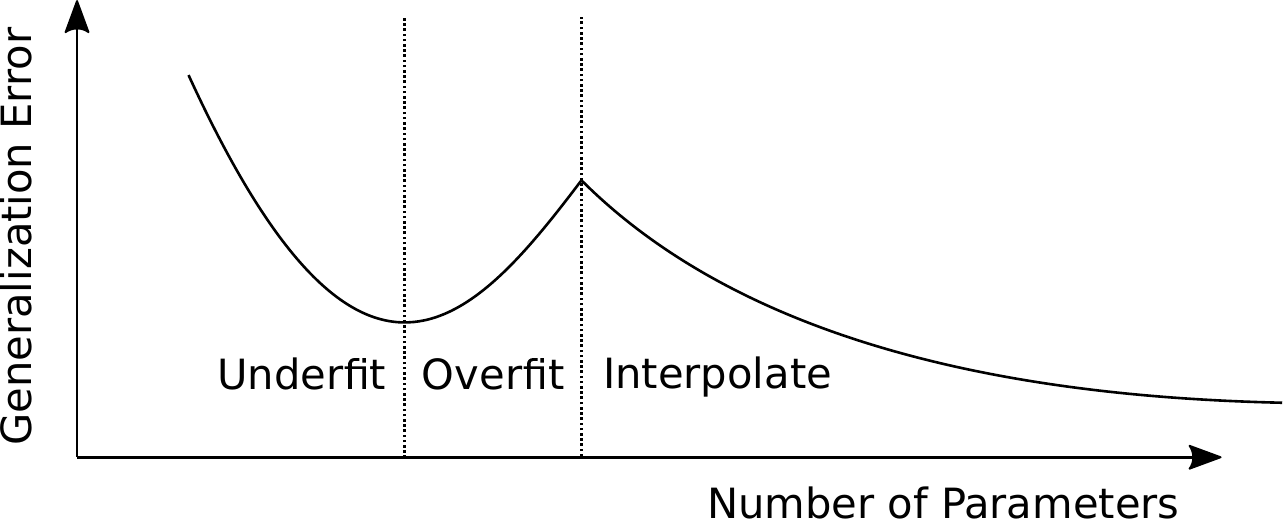}
   \end{center}
   \caption{Cartoon of the `double descent' generalization often observed in neural networks. }
\label{fig:double}
\end{figure}

Double descent generalization was observed in \cite{belkin2018reconciling}, has been confirmed 
in different architectures \cite{nakkiran2019deep} and is, in some settings, theoretically well understood \cite{hastie2019surprises}. 
It refers to a particular dependency of 
generalization error (or empirical error on a test set) on the number of free parameters 
in some machine-learning models. Namely, that with increasing number of parameters, 
the generalization error first goes down, then up and then down again (hence `double descent'). 
See Fig. \ref{fig:double}.
In models that have this behavior, the best performing models are 
the ones with the most parameters. These models are said to \emph{interpolate} the data. 

In this paper we point out that \emph{interpolating} FLOP-efficient ConvNets may be possible, by 
increasing their parameter density (per FLOP) using HyperNetworks. 

\subsection{HyperNetworks and Dynamic Networks}
HyperNetworks \cite{ha2016hypernetworks, schmidhuber1992learning} are neural networks in which some weights (or convolutional filters) 
are the outputs of a neural (sub-)network. More generally neural networks, whose parameters
 are the results of computations have been proposed in many different forms, e.g. \cite{stanley2002evolving, gomez2005evolving}. 
 Dynamic convolutions \cite{klein2015dynamic} and dynamic filter networks \cite{jia2016dynamic} propose predicting 
 the filters of a ConvNet or convolution layer from an input image or feature maps. \cite{DHP} uses HyperNetworks to prune (sparsify) a network,
 which is an alternative approach to using HyperNetworks to boost efficiency.

In this paper we apply such HyperNetworks (dynamic convolutions) and re-scale their fully-connected layers
to break the fixed parameter to FLOP ratio of ConvNets. 

\subsection{CondConv}
Conditional convolutions \cite{yang2019condconv} are a recent variant of dynamic filter networks, in which the
filter prediction has a particular form. Namely filters are predicted as weighted sums of 
several meta-filters; the weights in this weighted sum are predicted from feature-maps.
In this case the HyperNetwork is kept small to reduce computational burden.

In contrast to this, we deliberately increase the size of this HyperNetwork, so that 
parameter numbers increase. 

\subsection{Vision Transformer}
Recently it has been shown that transformer architectures can achieve excellent accuracy on
computer vision tasks \cite{dosovitskiy2020image}. Similar to the HyperNetwork-backed convolutions we use 
in this paper, transformers have a higher parameter to FLOP ratio than standard 
ConvNets. However, they have different inherent inductive biases (e.g. transformers
do not exhibit translation equivariance). 

\subsection{Efficient ConvNets}
There is a wide literature on designing efficient ConvNets. An excellent overview is given 
in \cite{deng2020model}. To our knowledge ours is the first paper pointing out the potential for reducing
FLOPs by decoupling FLOP-count form parameter-count in ConvNets. 

\section{Method}
\label{sec:method}
In this section we describe the basic HyperNetwork component that we will use repeatedly for our experiments.
We term this block a HyperConvolution. 

For an illustration see Fig.~\ref{fig:concept}, for a pseudo-code description Alg.~\ref{alg:HConv}, for code see the appendix. The functions in the pseudo-code are 
self-explanatory (modeled after pytorch functions), with the exception of `normalize', which ensures that the resulting filter's 
magnitude lie in a range that prevents fast activation explosion or decay at increasing depth;
see Sec.~\ref{sec:norm} for details. 

\begin{algorithm}
\caption{HyperConvolution}\label{alg:HConv}
\begin{algorithmic}[1]

\BState \textbf{function} \textsc{HyperConvolution} (I, F, n, $f_W$, $f_H$)
\State // I: forward feature maps, shape(I): $(N, C_\text{I}, H_\text{I}, W_\text{I})$
\State // F: filter feature maps, shape(F): $(N, C_\text{F}, H_\text{F}, W_\text{F})$
\State // n: size of MLP layers
\State // $f_W$, $f_H$: width and height of predicted filter
\State // \emph{returns}: $\text{I}_\text{out}$, shape($\text{I}_\text{out}$): $(N, C_\text{O}, H_\text{I}, W_\text{I})$
\State $\text{F}_\text{flat} \gets \text{globalMaxPool(F).flatten()}$
\State $f_\text{flat} \gets \text{MLP}_\text{n}(\text{F}_\text{flat})$
\State $f \gets f_\text{flat}$.reshape$(N\times C_I$,$C_O$,$f_W$,$f_H$)
\State $f \gets \text{normalize}(f) + f_\text{bias}$
\State $\text{I}_\text{grouped} \gets \text{I}$.reshape$(1, N \times  C_\text{I}, H_\text{I}, W_\text{I}$)
\State $\text{I}_\text{out} \gets \text{GroupConv} (\text{I}_\text{grouped}, f, \text{n}_\text{groups}=N, \text{pad=Same})$
\BState \textbf{return} $\text{I}_\text{out}$.reshape$(N, C_\text{O}, H_\text{I}, W_\text{I})$

\end{algorithmic}
\end{algorithm}

The HyperConvolution block takes two inputs: A `forward input' $I$
which it will convolve with some filters (and then output) and secondly a `filter input' $F$ from which filters
are computed (with which to convolve the forward input). The $F$ is max-pooled to size $1\times 1$,
and then fed into a multi-layer perceptron (MLP). The output layer of this MLP is sized such that
 it can be reshaped into the required convolutional filters. 

Note that the size of the hidden layers in the MLP impacts the number
of parameters of the HyperConvolution block, but the number of FLOPs required for it,
does not scale with the input image size. For a typically sized input image (hundreds to thousands of pixels
per dimension) and a hidden layer in the range of hundreds to thousands of neurons, the FLOPs required for
the MLP become negligible. Due to this, resizing the MLP allows increasing parameter count
at minor impact on FLOPs. 

Further note that each sample in the forward input batch is convolved with its own set of filters. In 
practice this means that the convolution is best implemented as a grouped convolution (see Alg.\ref{alg:HConv}).

In numbers, the ratio of FLOPs to parameters in a HyperConvolution block is
\begin{equation}
\frac{N_\text{FLOP}}{N_\text{Param}} = \frac{ 2 \cdot C_\text{in} \cdot C_\text{out} \cdot f_H \cdot f_W \cdot W \cdot H + N_\text{FLOP-MLP}}{C_\text{in} \cdot C_\text{out} \cdot f_H \cdot f_W + N_\text{Param-MLP}}
\end{equation}
Note that $N_\text{Param-MLP}$ and $ N_\text{FLOP-MLP}$ are independent of the input resolution and for an MLP without biases $N_\text{Param-MLP} / N_\text{FLOP-MLP} = 1 / 2$ (if we count one multiply-accumulate as 2 FLOPs).
In contrast for a standard convolution layer this is
\begin{eqnarray}
\frac{N_\text{FLOP}}{N_\text{Param}} &=& \frac{ 2 \cdot C_\text{in} \cdot C_\text{out} \cdot f_H \cdot f_W \cdot W \cdot H}{C_\text{in} \cdot C_\text{out} \cdot f_H \cdot f_W} \nonumber \\
	&=& 2 \cdot W \cdot H
	\label{eq:ratio}
\end{eqnarray}
In the standard convolution, clearly this ratio can only be changed, by modifying the input resolution. 

\subsection{Normalization of Predicted Filters}
\label{sec:norm}
In standard neural networks it has been observed that it is helpful for stable convergence of training to initialize 
weight matrices (or convolution filters) such that the variance of activations propagating through the network does not 
grow or decay too quickly \cite{glorot2010understanding}. For our HyperConvolution layers we ensure this by normalizing the output of the MLPs given in 
the HyperConvolution
Alg. \ref{alg:HConv} (the notation in the following is consistent with that algorithm). 

We assume the reshaped output $f$ of the MLP is normally distributed 
\begin{equation}
f \sim \mathcal{N}(0,1)
\end{equation}
Aiming for `He initialization' \cite{he2016deep} we need to scale this with $\sqrt{2/\text{fan-in}}=\sqrt{2/(C_\text{in} \cdot f_W \cdot f_H)}$.
We can also achieve the desired variance by the  following normalization:
\begin{equation}
f \gets f \cdot \frac{\sqrt{2 \cdot f_W \cdot f_H} }{\sqrt{C_\text{in} \cdot \pi / 2} \cdot |f|.\text{sum}([2,3])  }
\label{eq:norm}
\end{equation}
where we use that the mean of the half-normal distribution is $\sigma \sqrt{2 / \pi}$ such that
\begin{equation}
\mathbb{E}\left[|f|.\text{sum}([2,3])\right] = \sqrt{2 / \pi} \cdot f_W \cdot f_H
\end{equation}
The normalization of Eq. \ref{eq:norm} further fixes the L1-norm of each filter to 1. This 
has an effect similar to Instance-Normalization, in that output channels have approximately 
the same magnitude.

\subsection{Memory Requirements and Parameters}
For standard convolution layers increasing the number of parameters, will increase 
the memory requirements (RAM usage) of the network. The dominating contributor
to this are the feature maps (not the filters) in most standard architectures.

When using HyperConvolutions, two networks with the same number of parameters can have very different 
memory requirements.  
Consider as an example one network with many channels per HyperConvolution
and few hidden units in the MLP, and another with few channels and many hidden units
in the MLP. 

The networks we use in this paper have comparatively low memory requirements,
although they have many parameters (see Tab. \ref{tab:bigtab}).
This is due to the fact that 
 our networks have comparatively fewer feature maps (or channels per convolution).

\subsection{HyperConvolution as a Non-local Method}
A different motivation for the use of HyperConvolutions is that they
incorporate global information into the high-resolution pathway of a 
ConvNet through the HyperNetwork that predicts convolution filters.
Non-local information is well-known to be useful in deep learning
approaches to image processing \cite{wang2018non}.

Furthermore, a Hyper-Convolution layer bears some similarity to `classical' (non-deep-learning)
 non-local methods: Both BM3d \cite{dabov2006image} and a Hyper-Convolution
layer with sigmoid activation function perform the following operation
on an abstract level: From an input image $I$ feature patches $P$ are computed 
followed by a similarity between subsections
of $I$ and $P$. Here the commonality ends. In our approach, the patches $P$  
are the predicted convolution filters and the similarity is the sigmoid of a dot-product, where for BM3d
$P$ are image sub-blocks and the similarity is a carefully
crafted block-distance measure. Nevertheless this additionally motivates the use of Hyper-Convolutions from a 
perspective of inductive biases.

\begin{figure}[th]
\begin{center}
	\includegraphics[width=\linewidth]{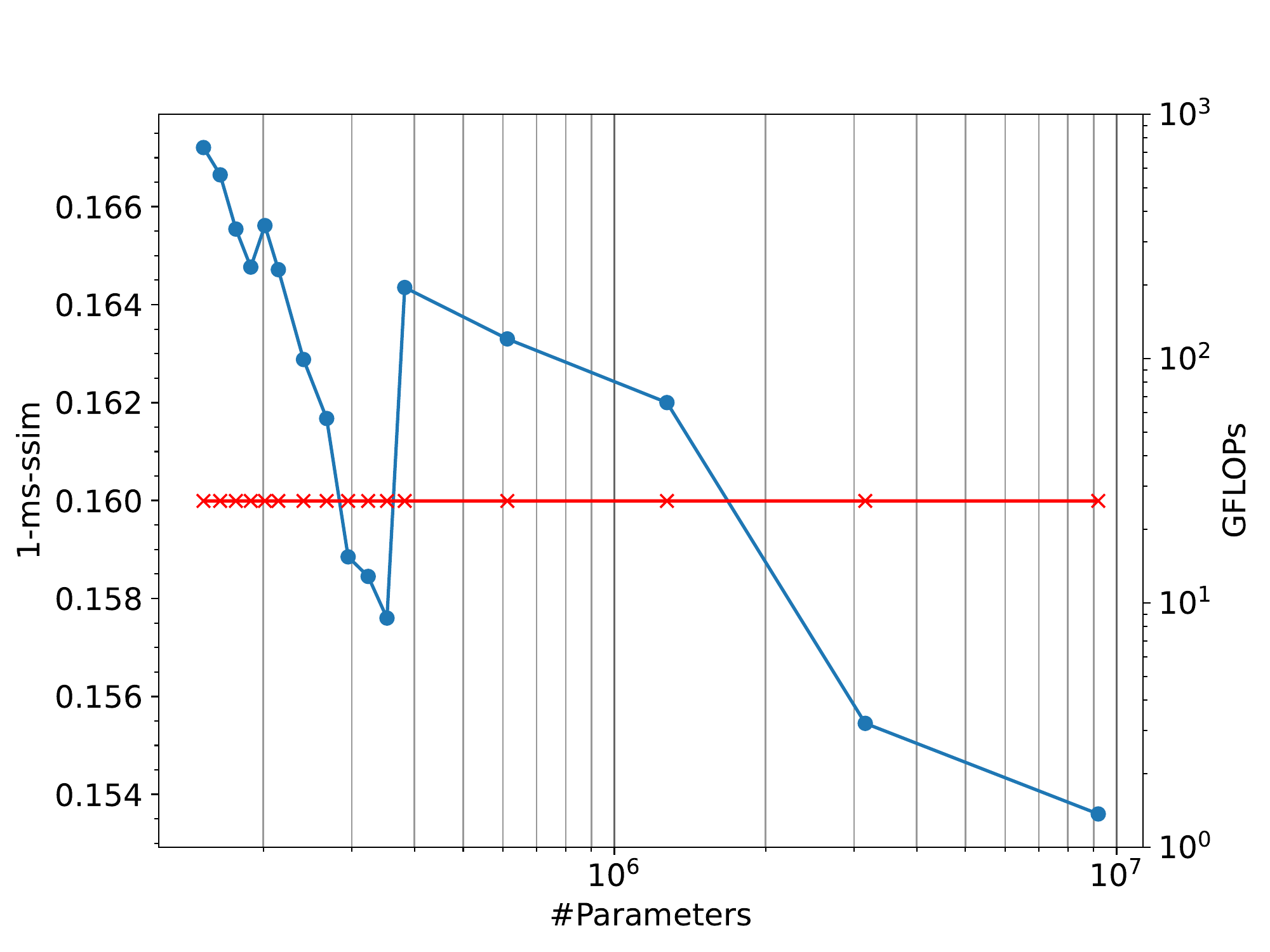}
\end{center}
\caption{Double Descent curve for our U-Net with HyperConvolutions: Interpolating generalization at constant FLOPs by increasing MLP size.}
\label{fig:mainresult}
\end{figure}

\begin{figure}[th]
\begin{center}
\includegraphics[width=\linewidth]{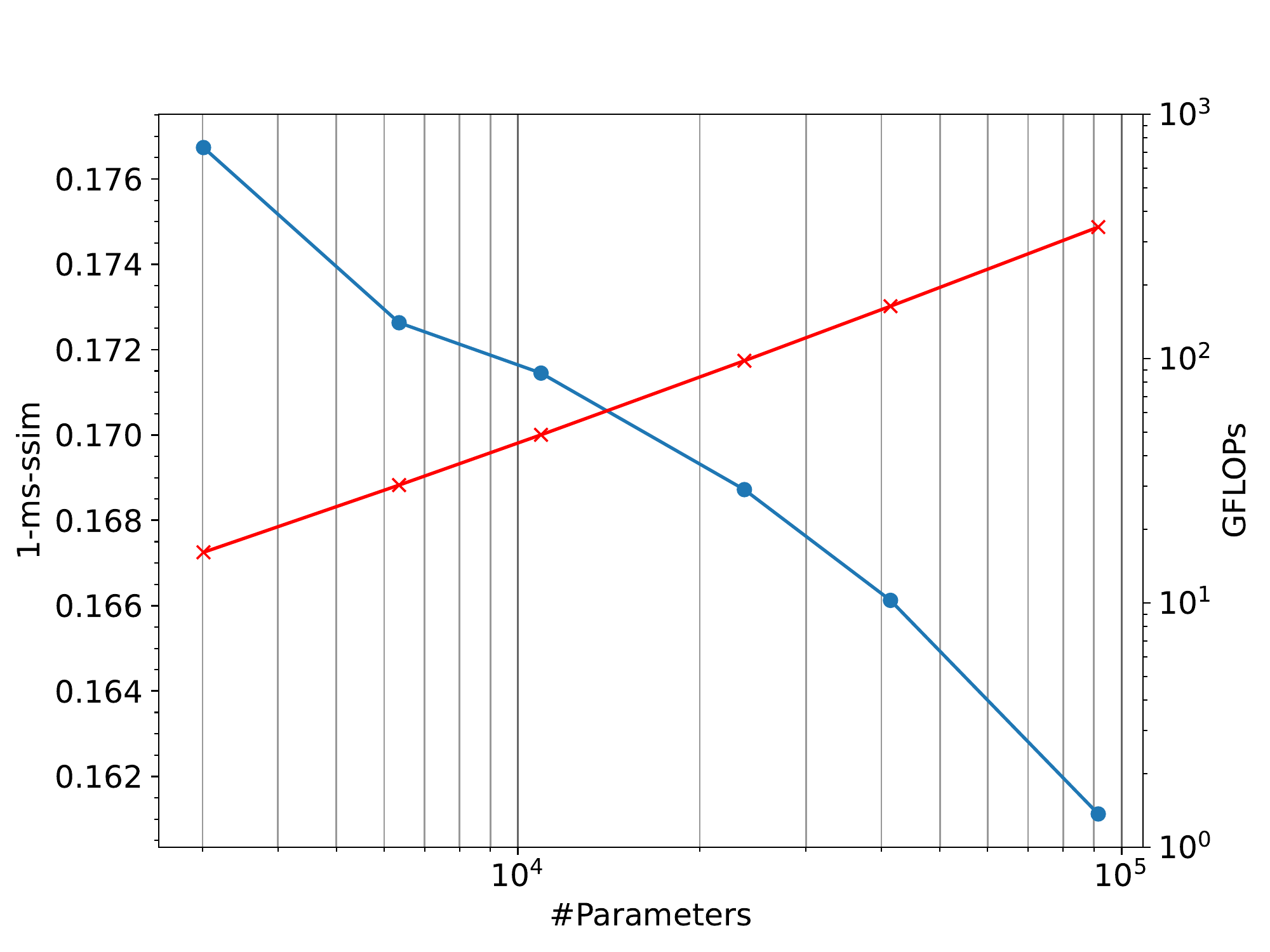}
\end{center}
\caption{Corresponding curve for the same U-Net with standard convolutions. FLOPs grow very large before curve flattens.}
\label{fig:mainresultb}
\end{figure}

\begin{table*}
\centering
\begin{tabular}{|l||r|r|r|r|r|r|r|}
\hline
Network ($n_\text{fwd}, n_\text{embed}, n_\text{hid}$) & FLOPs & Parameters & CPU time & Max. Conv. Mem. & PSNR & MS-SSIM & SSIM \\
\hline
\hline
PyNet \cite{ignatov2020replacing} & 43 T & 47 M & 120 s & 29.7 Gb & 21.19 & \textbf{0.8620} & - \\
PyNet-CA \cite{kim2020pynet} & 45 T & 51 M & 131 s & 32.8 Gb & 21.22 & 0.8549 & 0.7360 \\
AWNet 4-channel \cite{dai2020awnet} & 9.4 T & 52 M & 55 s & 27.8 Gb & \textbf{\color{red} 21.38} & 0.8590 & 0.7451 \\
SPADE \cite{park2019semantic} * &  0.8 T* & 97 M* & -& - & 20.96 & 0.8586 & - \\
DPED \cite{ignatov2017dslr} * &  1.4 T*  & 4 M* & -& - & 20.67  & 0.8560 & - \\
UNet \cite{ronneberger2015u} * &  3.9 T* & 17 M* & - & - & 20.81 & 0.8545 & - \\
\hline
\hline
Ours (96, n/a, n/a) no HC & 1.3 T & 0.4 M & 11 s & 3.8 Gb & 19.93 & 0.8463 & 0.7213 \\
Ours (64, n/a, n/a) no HC & 0.6 T & 0.2 M & 7 s & 2.6 Gb & 19.82 & 0.8446 & 0.7185 \\
\hline
Ours (64, 32, 2048) & 0.7 T & 276 M & 12 s & 3.4 Gb & \textbf{21.37} & \textbf{\color{red} 0.8640} & \textbf{\color{red} 0.7509} \\
Ours (32, 32, 2048) & 0.3 T & 95 M & 6 s & 2.2 Gb & 21.11 & 0.8618 & \textbf{0.7466} \\
Ours (32, 16, 2048) & 0.2 T & 90 M & 5 s & 1.8 Gb & 21.15 & 0.8617 & 0.7471 \\
Ours (8, 32, 4000) & 0.1 T & 113 M & 3 s & 1.3 Gb & 20.22 & 0.8428 & 0.7232 \\

\hline
\end{tabular}
\caption{Performance comparison on ZRR \cite{ignatov2020replacing} of the proposed networks and state-of-the-art (single networks, w/o ensembling). FLOPs are computed assuming a 12.6Mpix input.
No HC indicates ablation experiments with standard convolutions in place of the HyperConvolution.
The first four columns are only given, where we had access to a pre-trained model. The two best results are marked in bold, the best in red. *Results reproted in \cite{ignatov2020replacing}, FLOPs and parameters estimated by hand.}
\label{tab:bigtab}
\end{table*}

\begin{table*}[t!]
\centering
\begin{tabular}{|l||r|r|r|r|r|r|}
\hline
Network & FLOPS & Param.s & CPU time & Max. Conv. Mem. & PSNR (valid / test) & SSIM (valid / test) \\
\hline
\hline
VDN  & 9.5 T &  \textbf{7.8 M} & 3.1 s / MPix & 2.3 GB & 39.36 / 39.26 & 0.917 / 0.955\\
\hline
Ours ($n_\text{channel} / 3$) & \textbf{1.4 T} & 55.0 M &  \textbf{1.5 s / MPix} & \textbf{1.0 GB} & 39.40 / 39.23 & \textbf{0.918 / 0.957}  \\
Ours ($n_\text{channel} / 2$) & 2.9 T & 119.6 M &  2.5 s / MPix & 1.4 GB & \textbf{39.42 / 39.27} & \textbf{0.918 / 0.957}\\

\hline
\end{tabular}
\caption{Comparison on SIDD \cite{abdelhamed2018high} of the original VDN \cite{yue2019variational} and the same network using our HyperConvolutions with fewer channels.}
 \label{tab:sidd}
\end{table*}

\begin{figure*}
\begin{center}
\subfloat{
\includegraphics[width=0.24\linewidth]{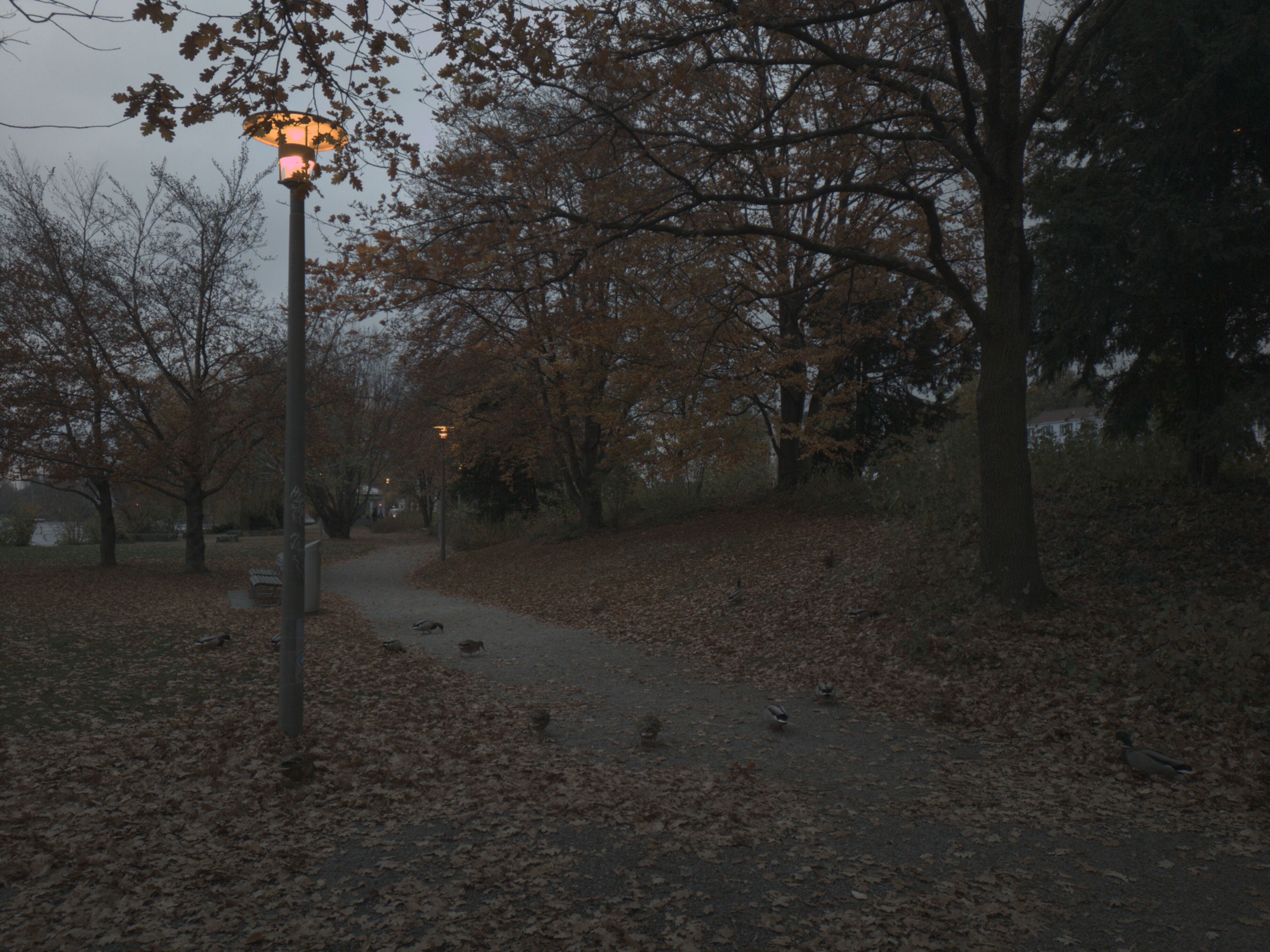}}
\hfill
\subfloat{
\includegraphics[width=0.24\linewidth]{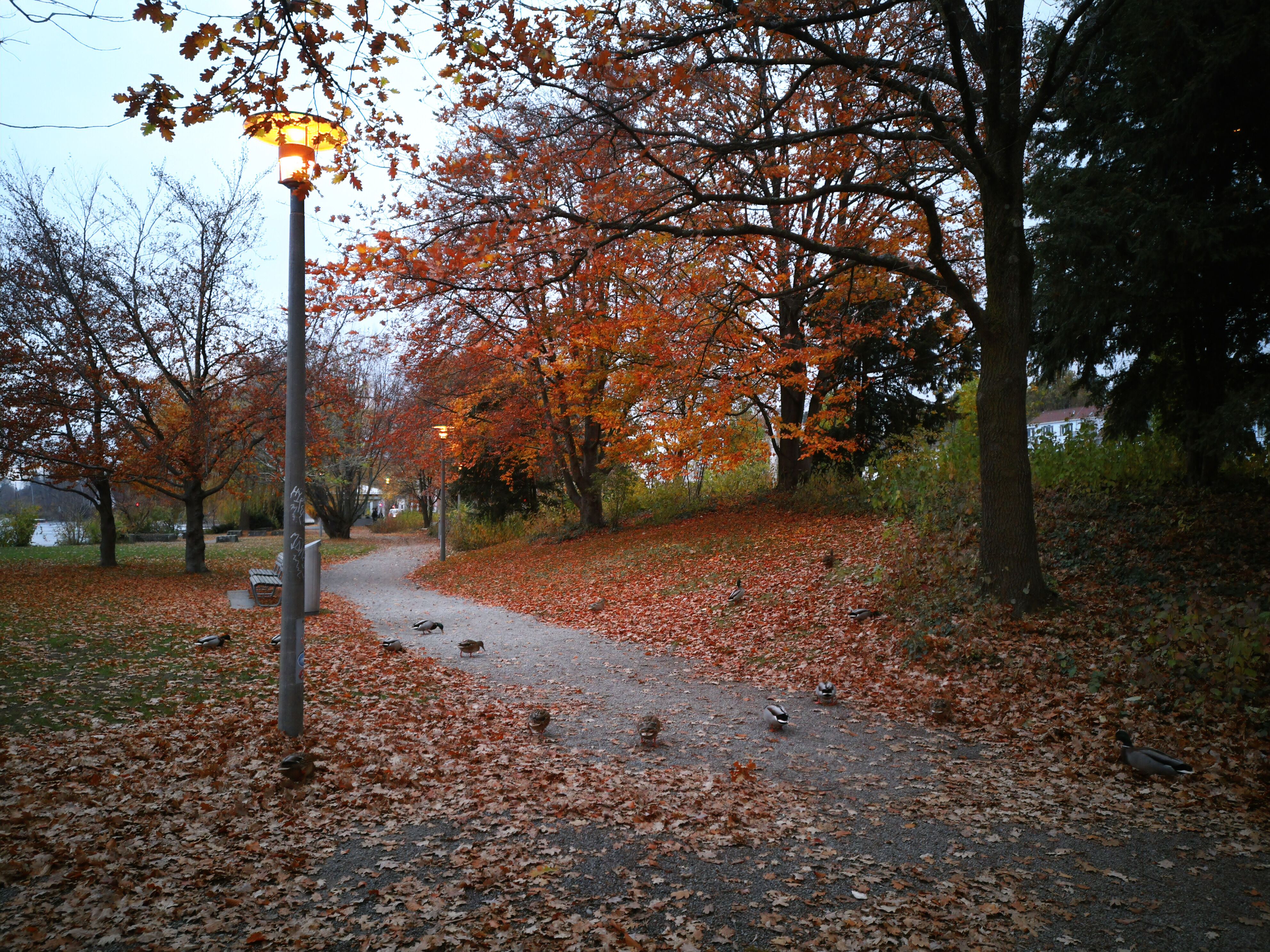}}
\hfill
\subfloat{
\includegraphics[width=0.24\linewidth]{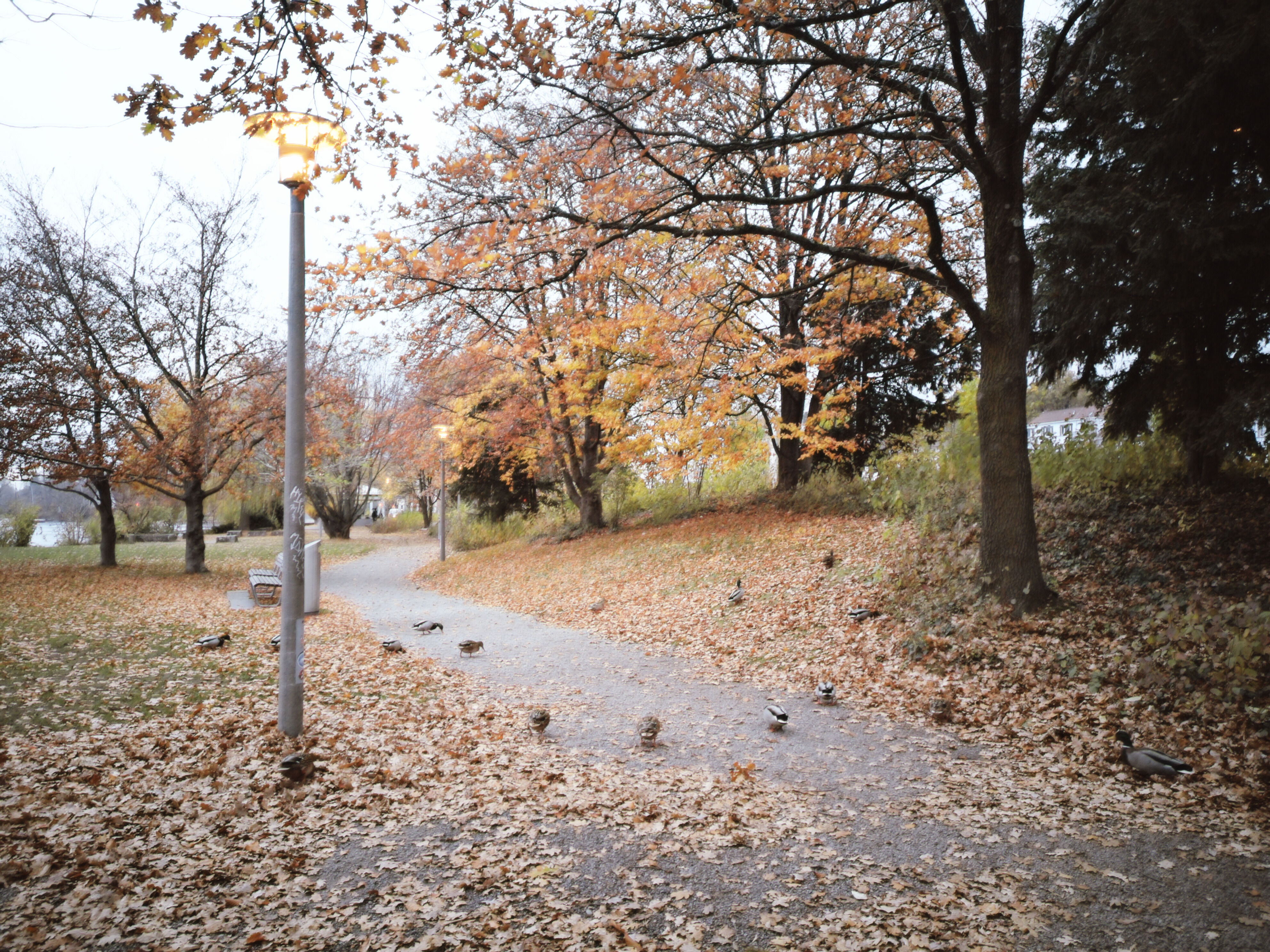}}
\hfill
\subfloat{
\includegraphics[width=0.24\linewidth]{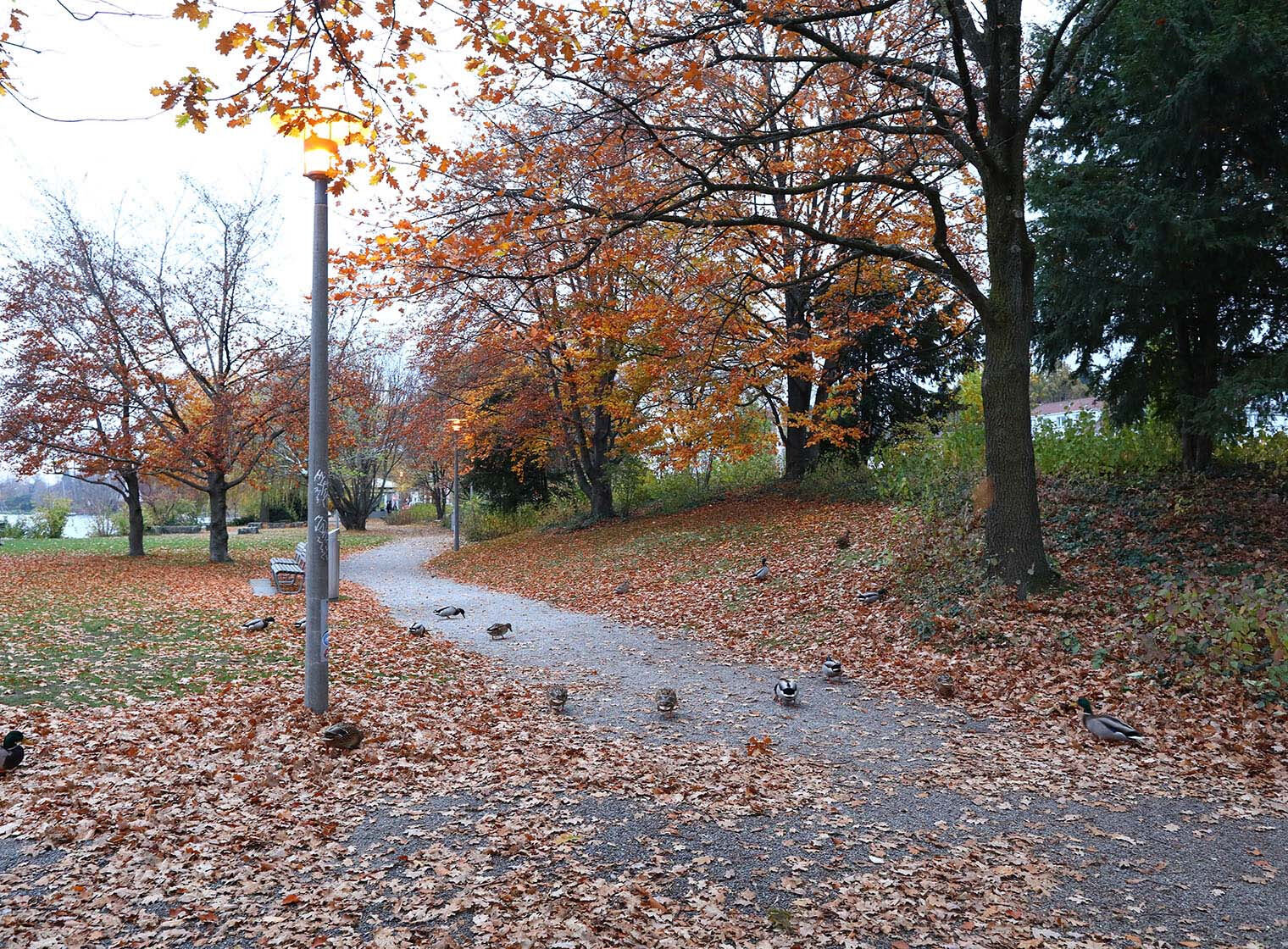}} \\ \vspace{-.3cm}

\subfloat{
\includegraphics[width=0.24\linewidth]{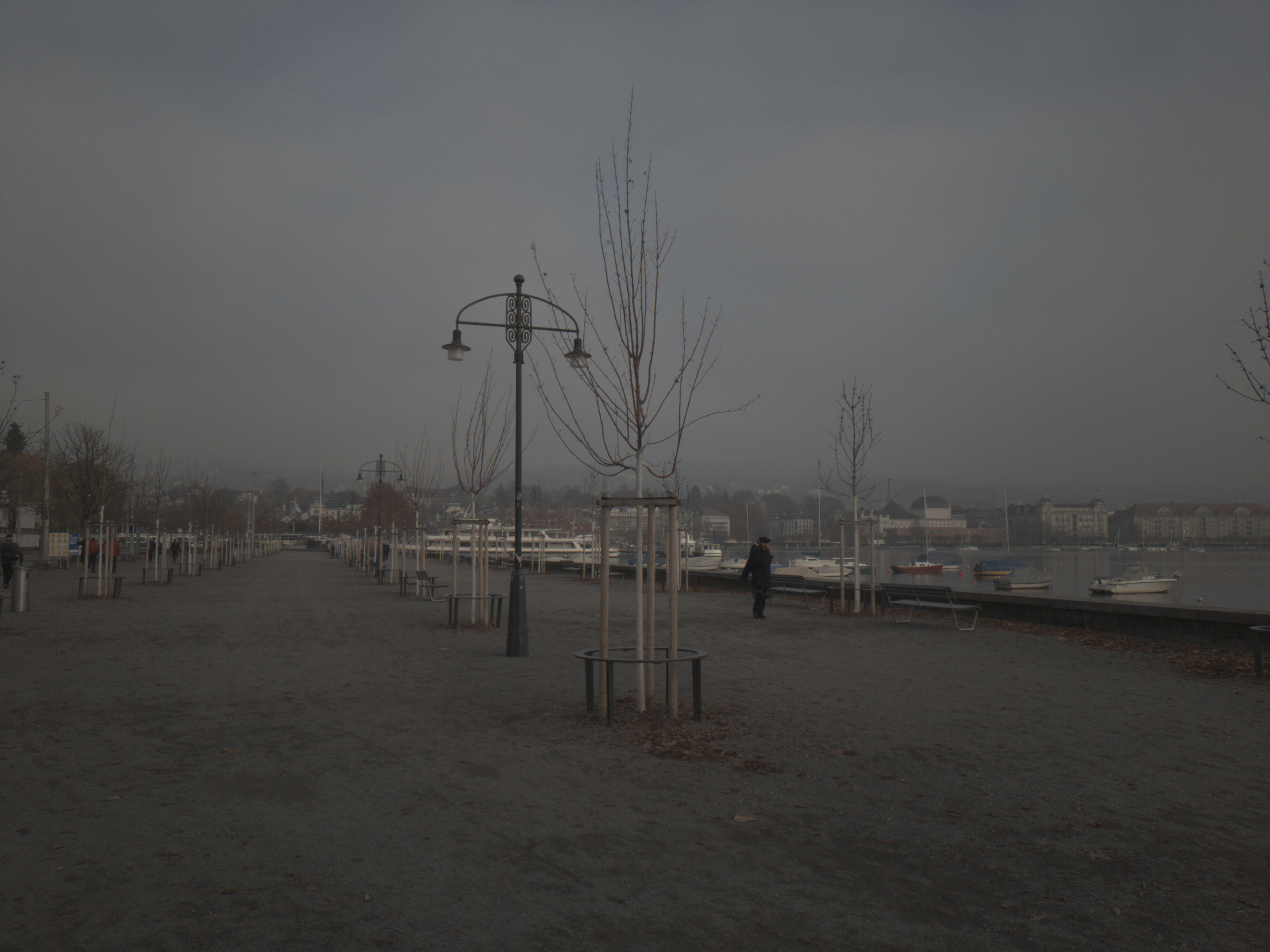}}
\hfill
\subfloat{
\includegraphics[width=0.24\linewidth]{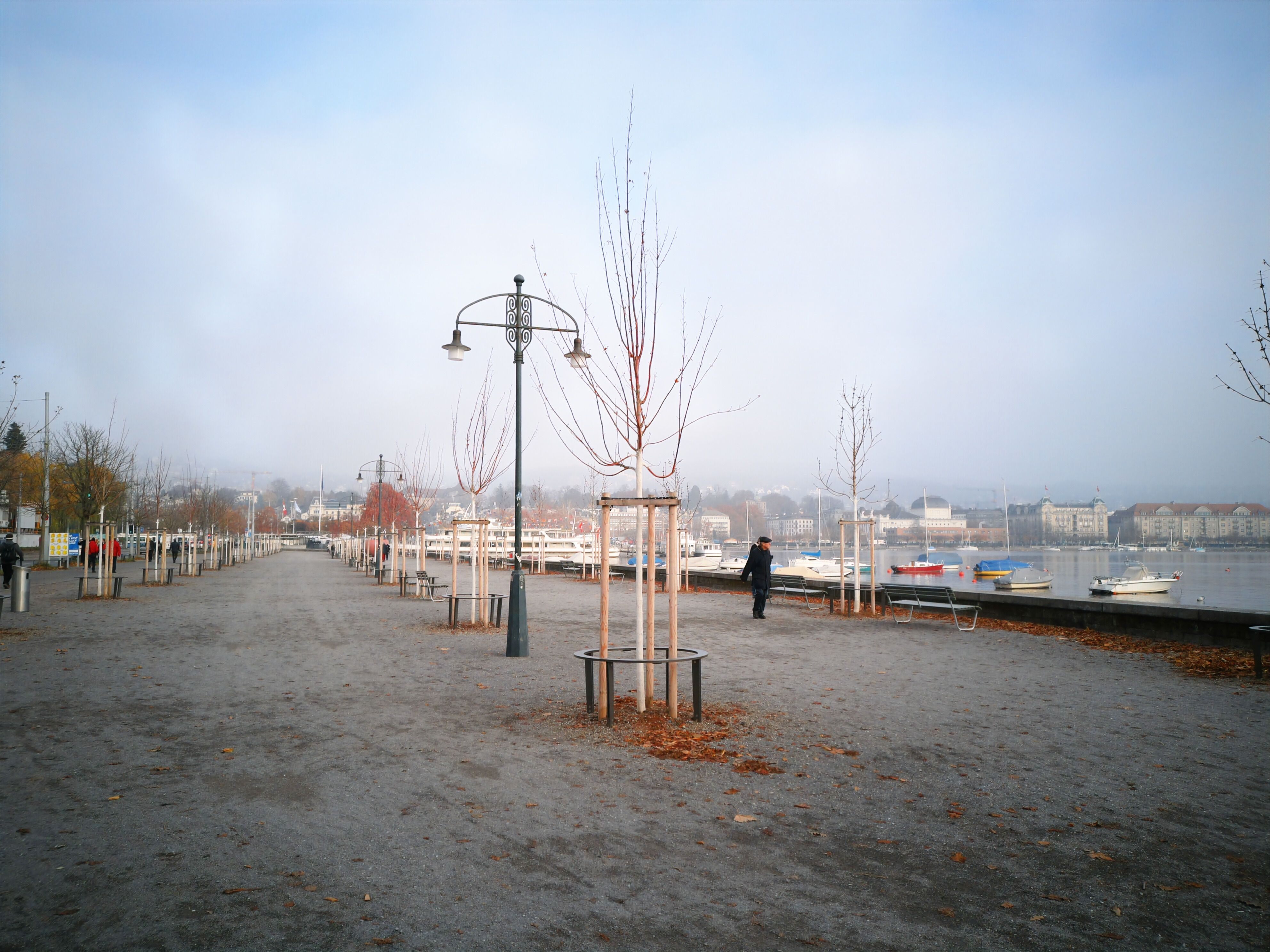}}
\hfill
\subfloat{
\includegraphics[width=0.24\linewidth]{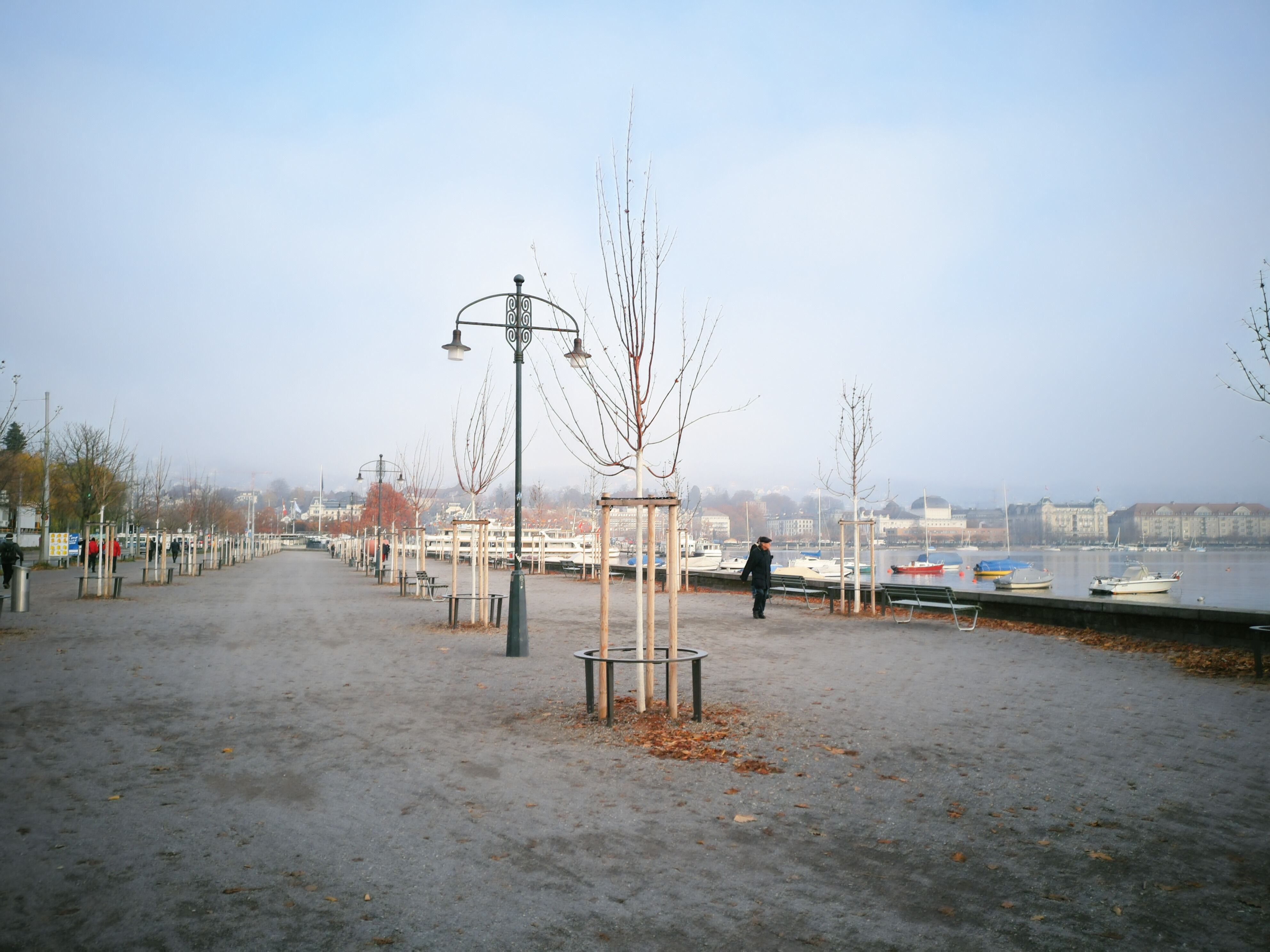}}
\hfill
\subfloat{
\includegraphics[width=0.24\linewidth]{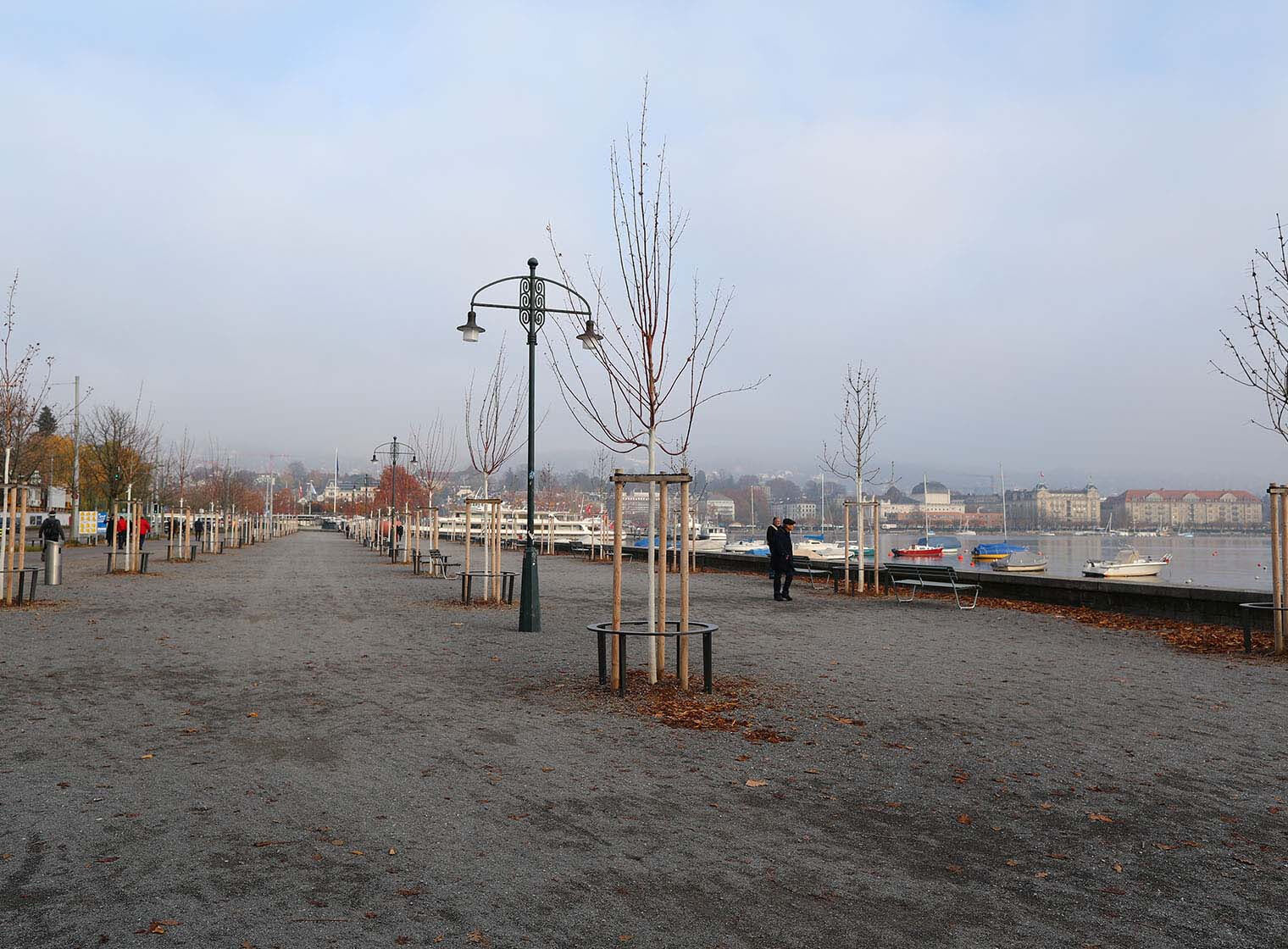}} \\ \vspace{-.3cm}

\subfloat{
\includegraphics[width=0.24\linewidth]{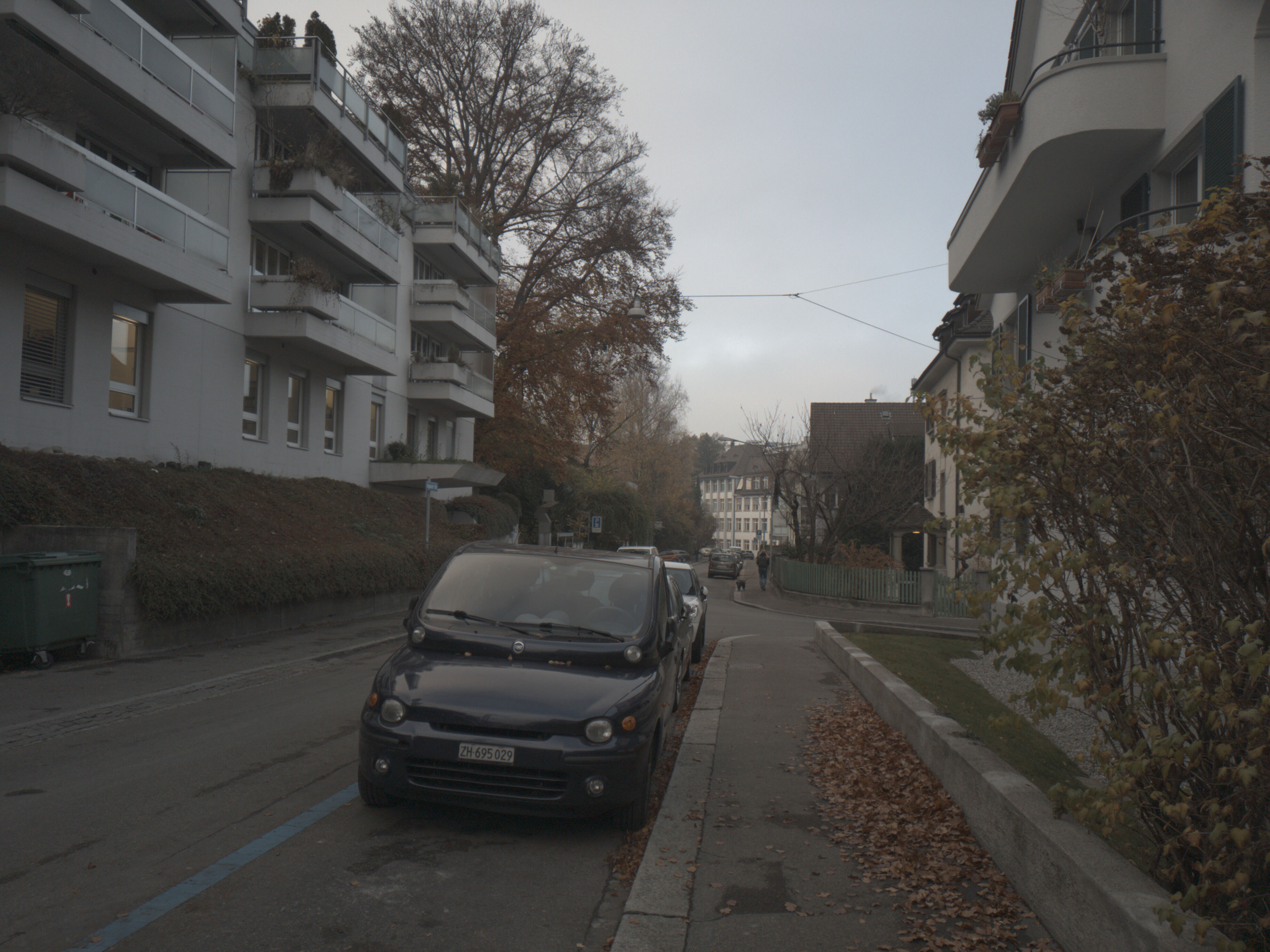}}
\hfill
\subfloat{
\includegraphics[width=0.24\linewidth]{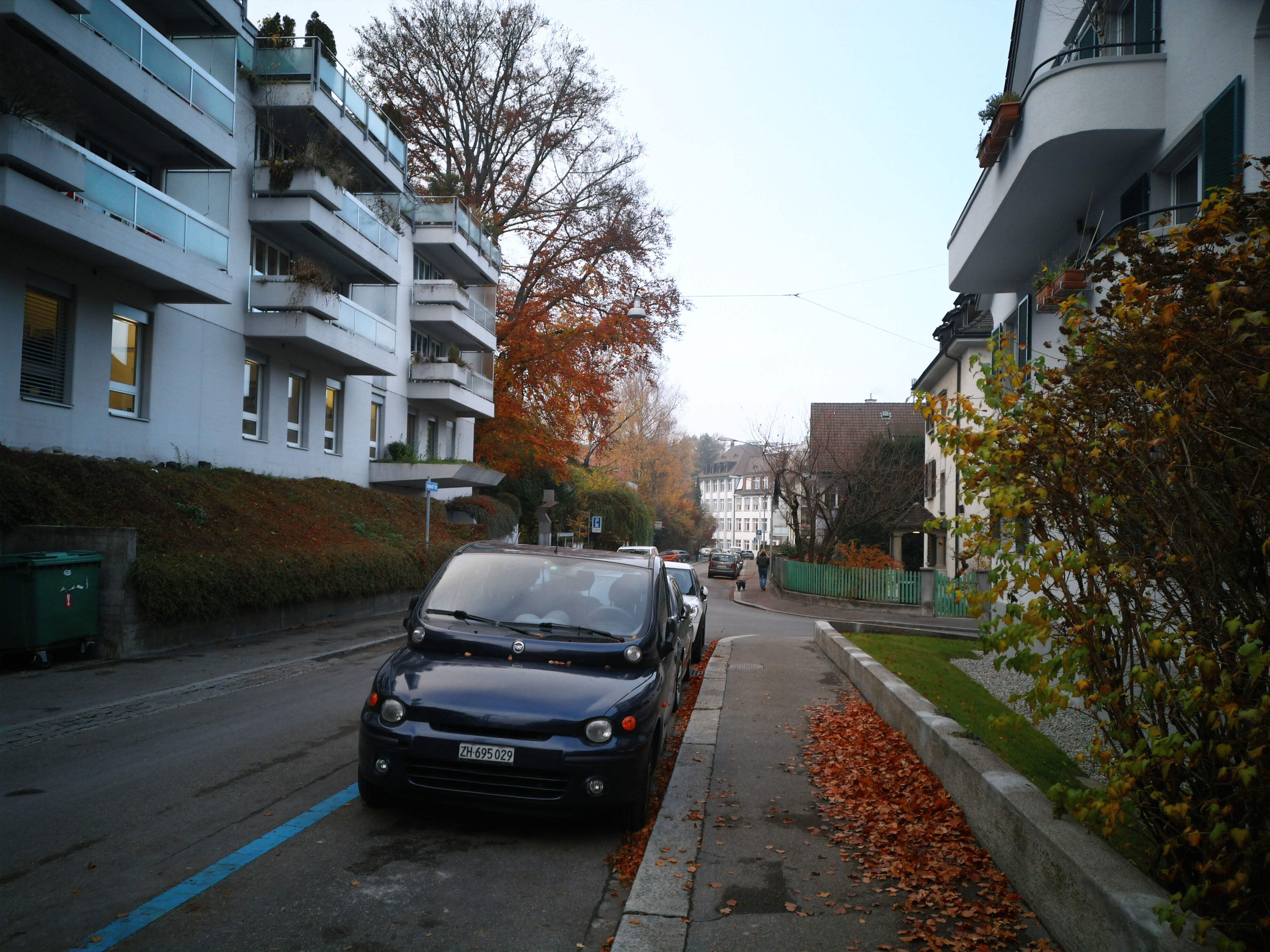}}
\hfill
\subfloat{
\includegraphics[width=0.24\linewidth]{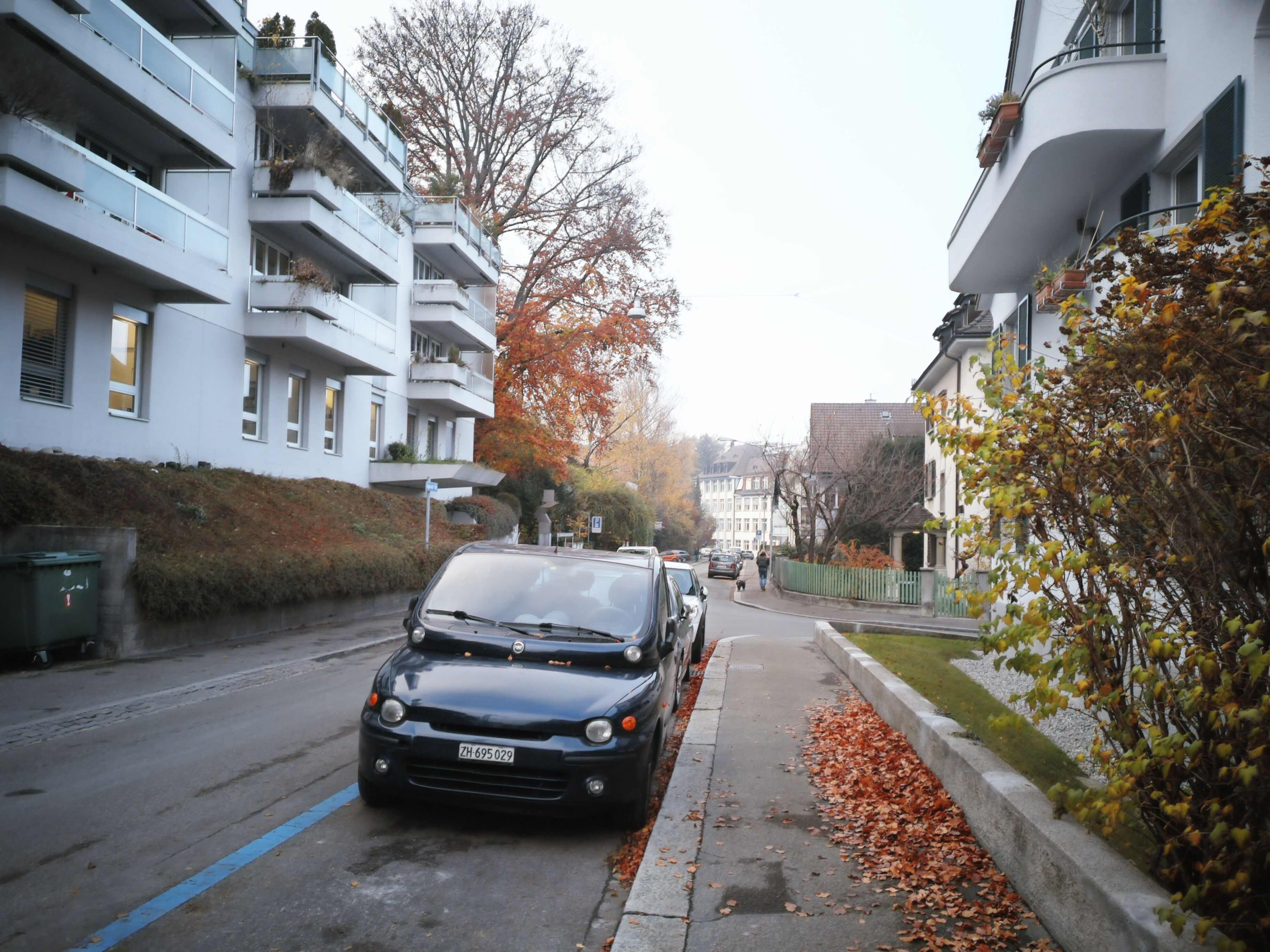}}
\hfill
\subfloat{
\includegraphics[width=0.24\linewidth]{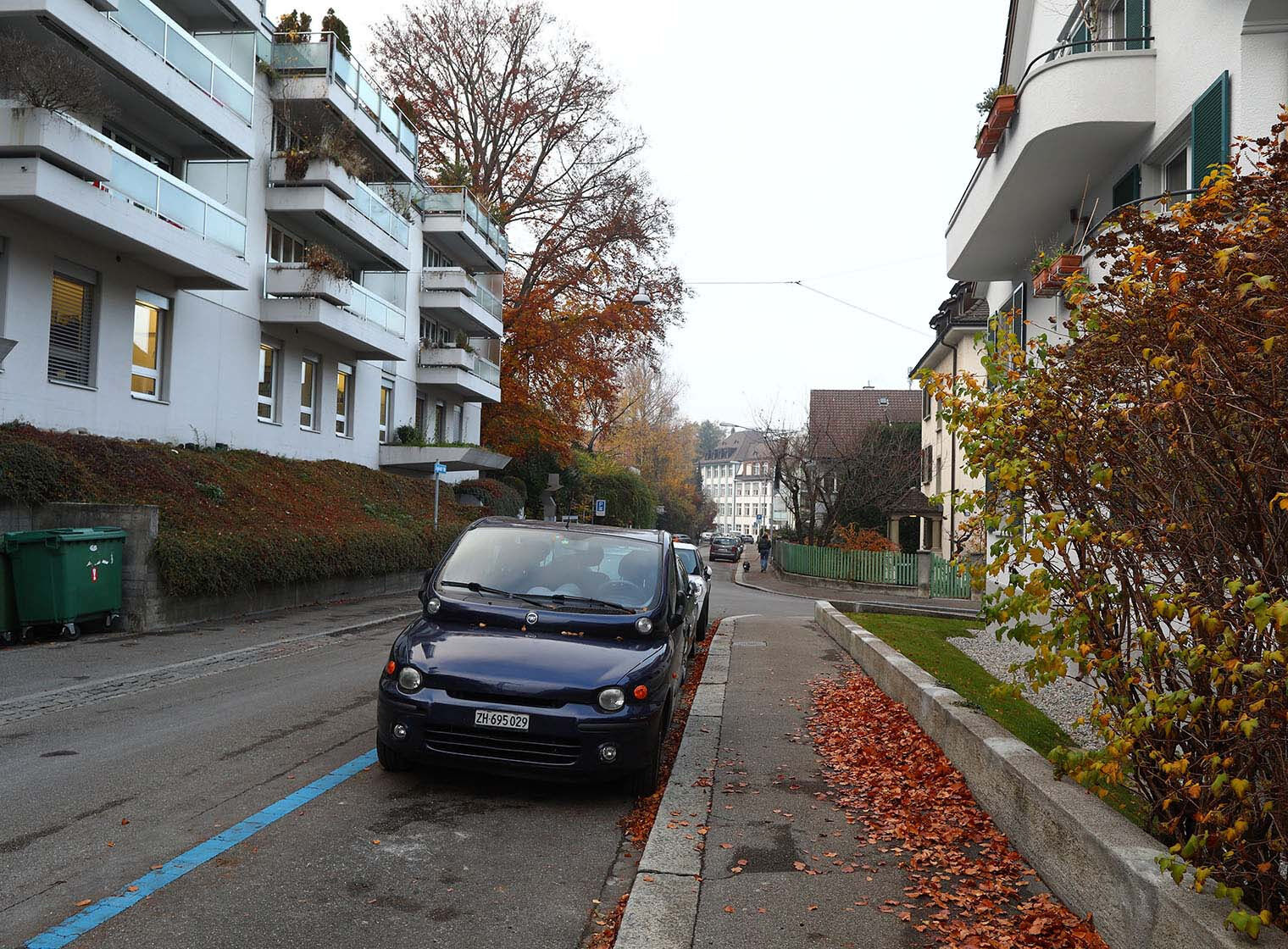}} \\ \vspace{-.3cm}

\subfloat[RAW visualized]{
\includegraphics[width=0.24\linewidth]{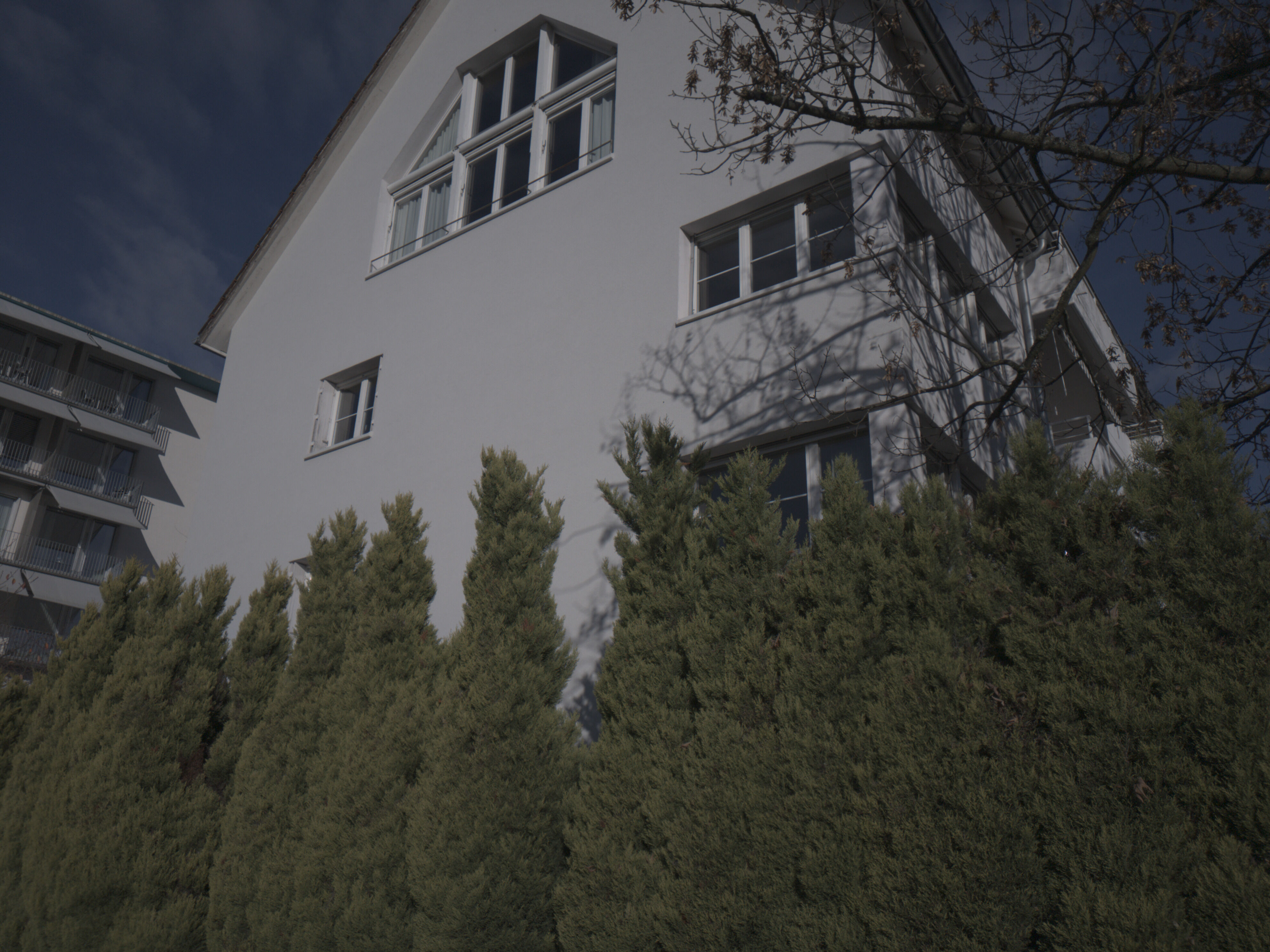}}
\hfill
\subfloat[AWNet]{
\includegraphics[width=0.24\linewidth]{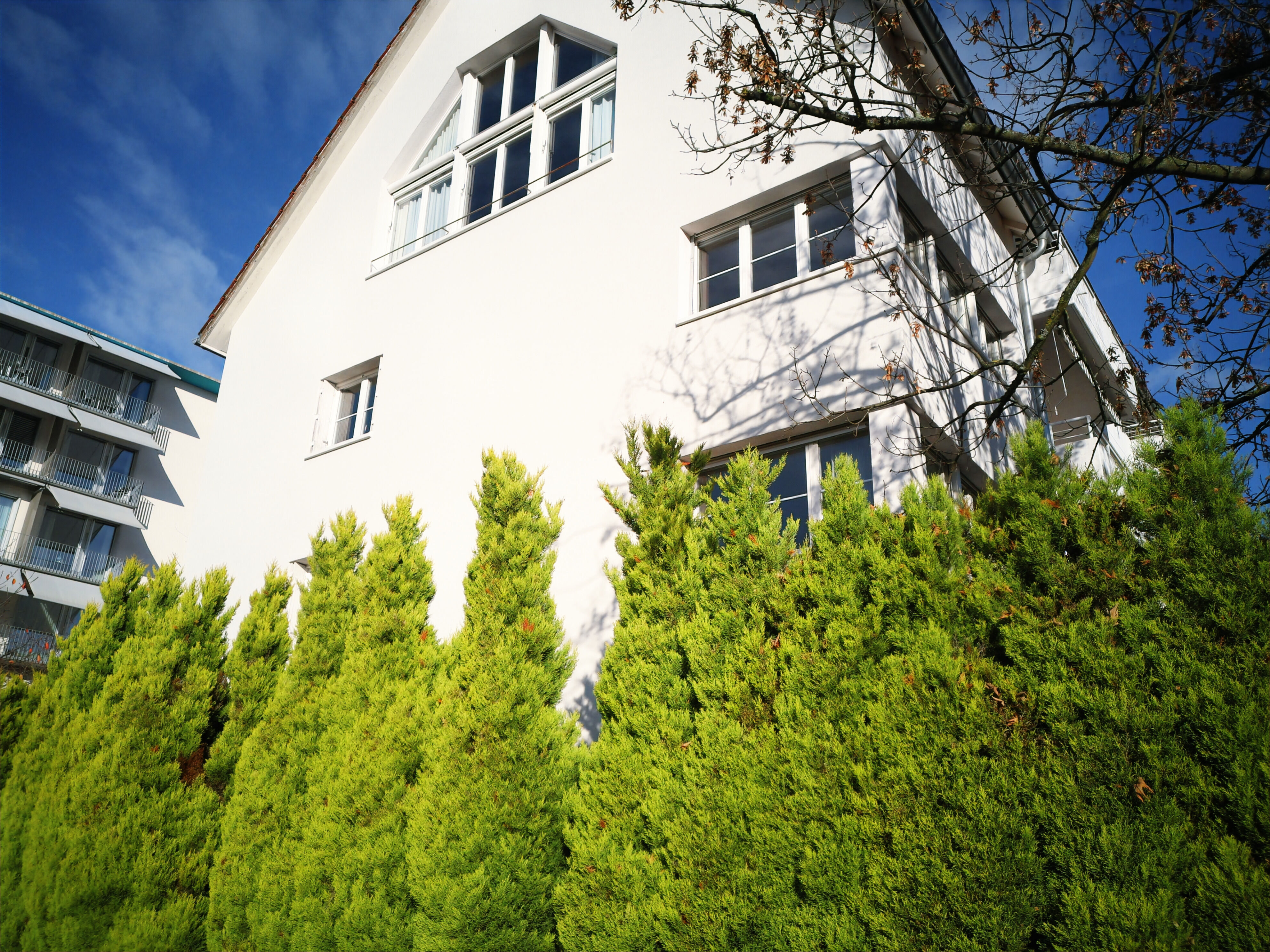}}
\hfill
\subfloat[Ours]{
\includegraphics[width=0.24\linewidth]{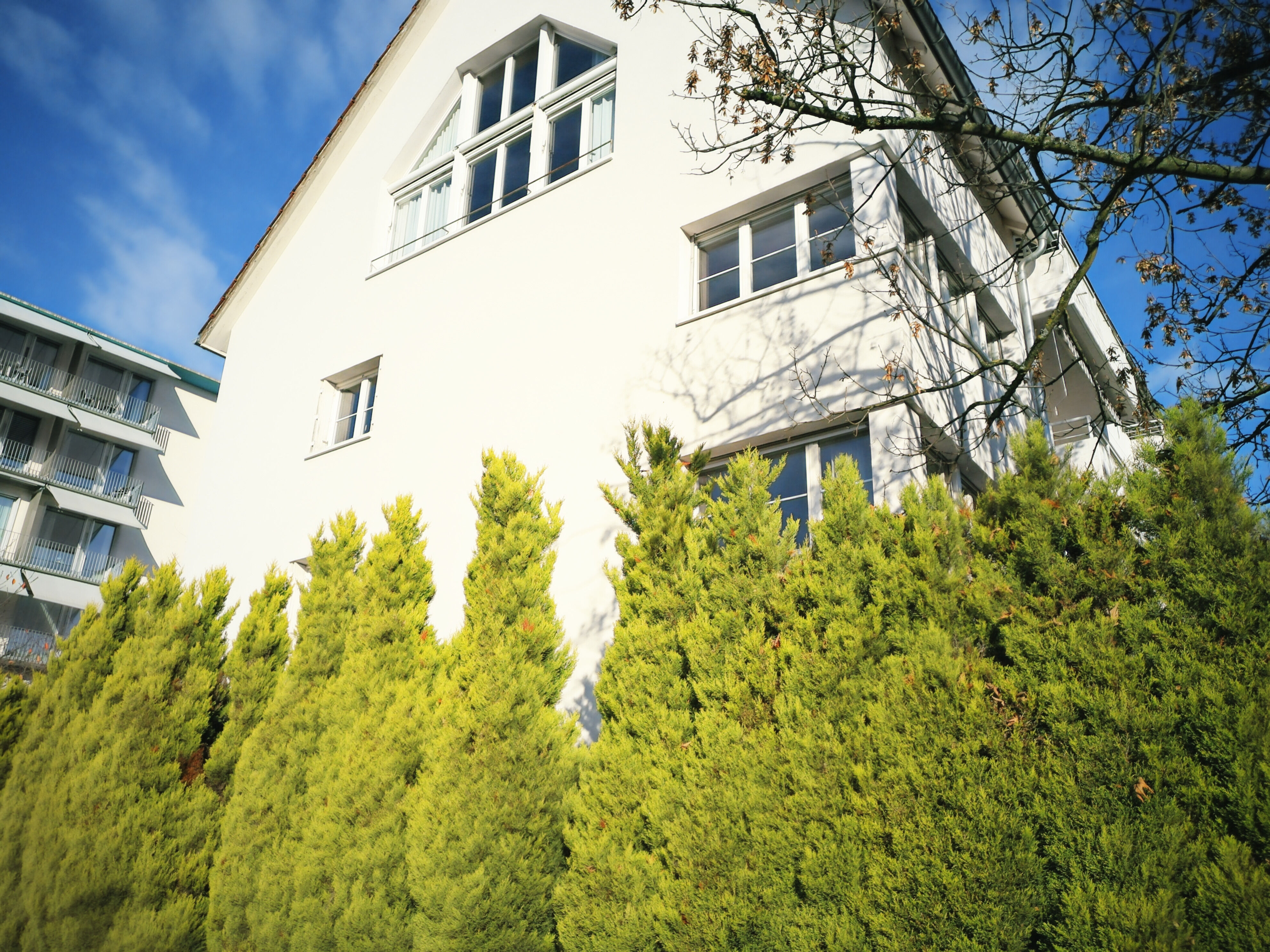}}
\hfill
\subfloat[DSLR (target)]{
\includegraphics[width=0.24\linewidth]{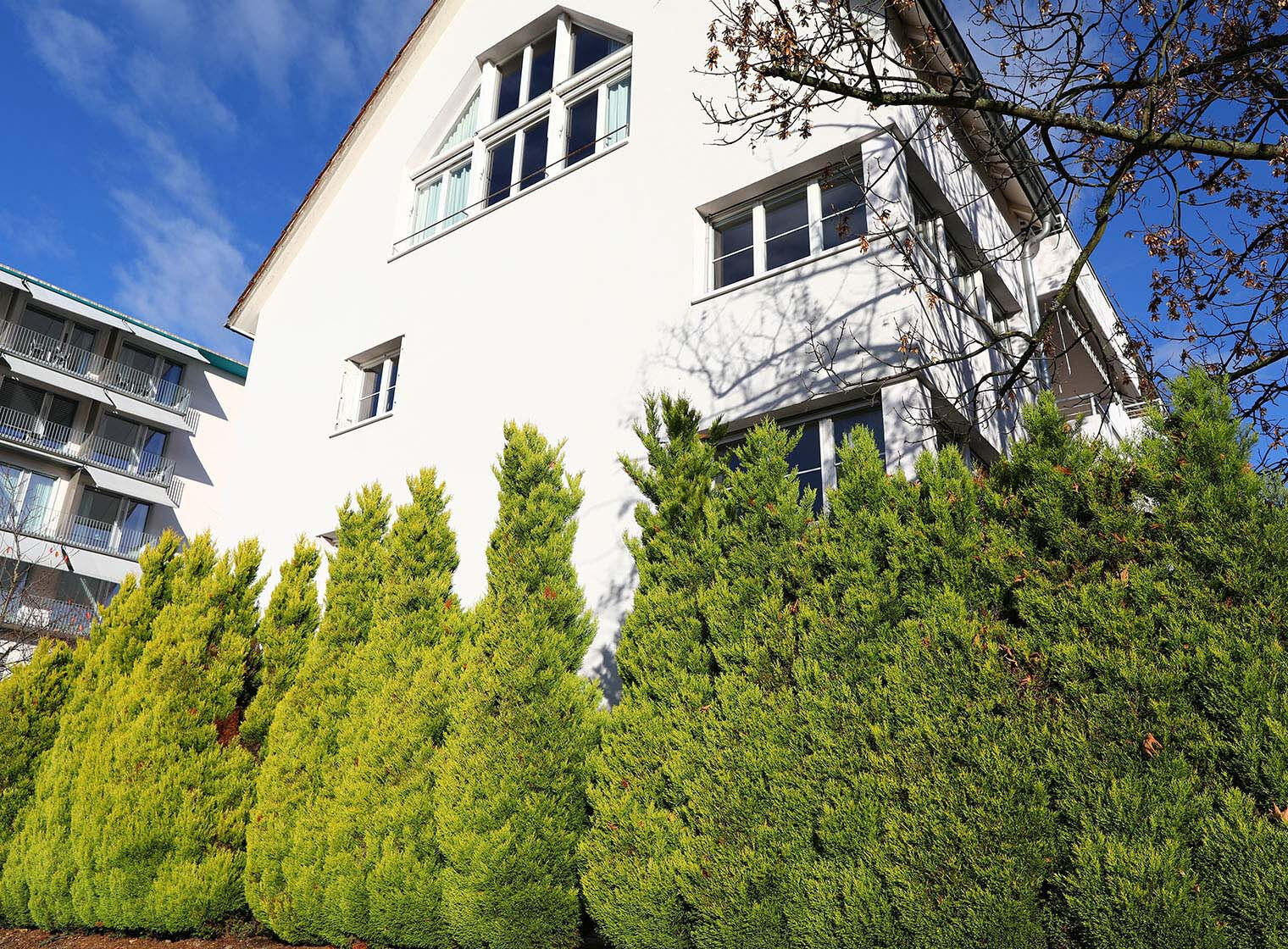}} 

\end{center}

   \caption{Comparison full images ZRR}
\label{fig:pics}
\end{figure*}

\begin{figure}
\begin{center}
\subfloat{
\includegraphics[width=0.185\linewidth]{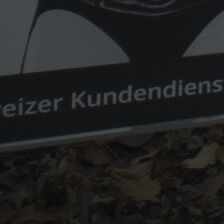}}
\hfill
\subfloat{
\includegraphics[width=0.185\linewidth]{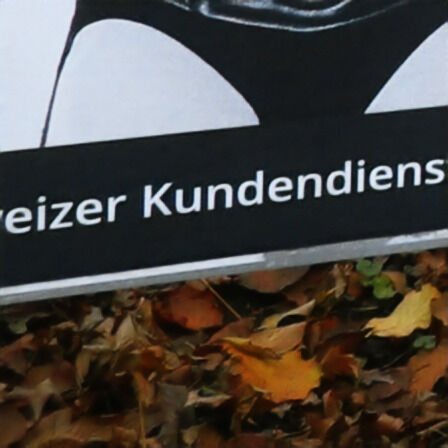}}
\hfill
\subfloat{
\includegraphics[width=0.185\linewidth]{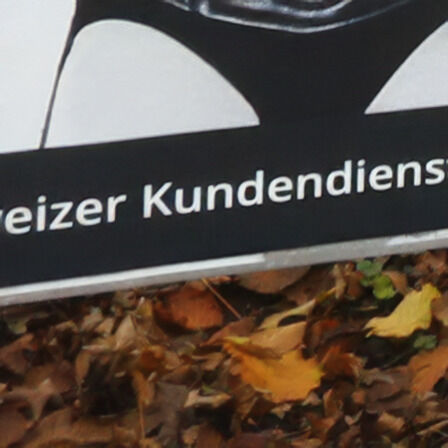}}
\hfill
\subfloat{
\includegraphics[width=0.185\linewidth]{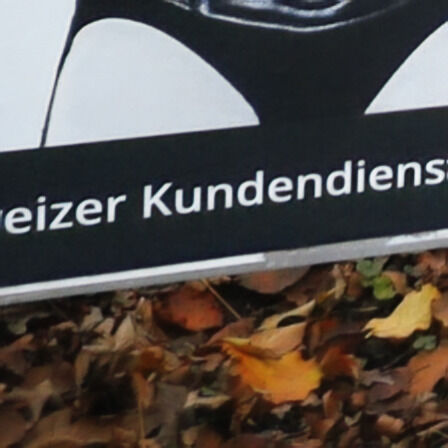}}
\hfill
\subfloat{
\includegraphics[width=0.185\linewidth]{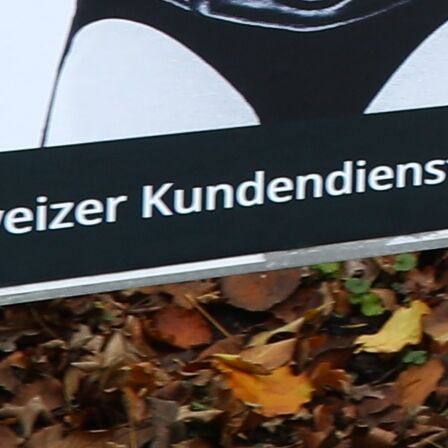}} \\ \vspace{-.3cm}

\subfloat{
\includegraphics[width=0.185\linewidth]{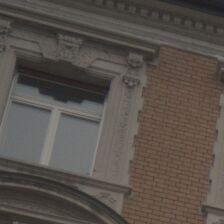}}
\hfill
\subfloat{
\includegraphics[width=0.185\linewidth]{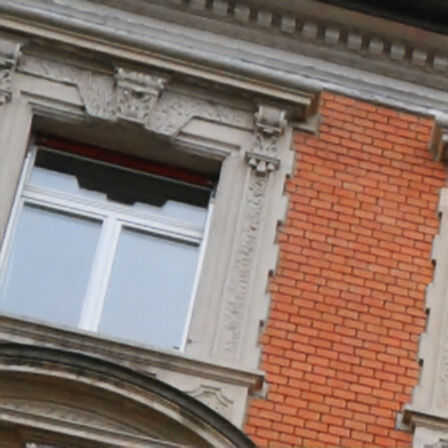}}
\hfill
\subfloat{
\includegraphics[width=0.185\linewidth]{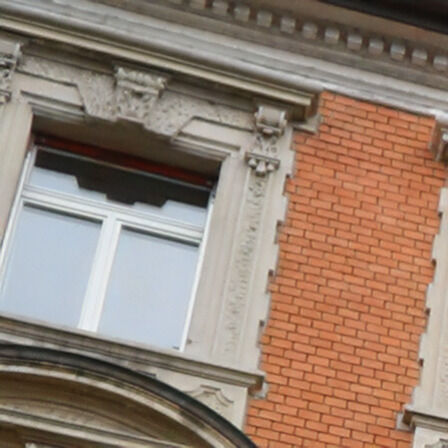}}
\hfill
\subfloat{
\includegraphics[width=0.185\linewidth]{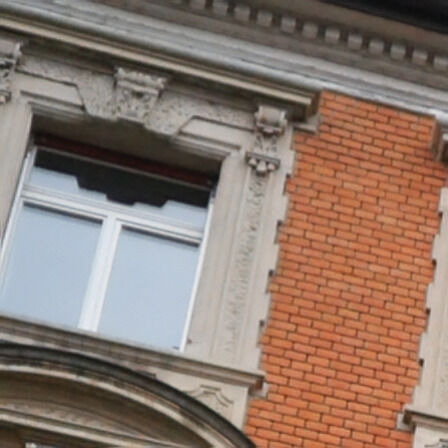}}
\hfill
\subfloat{
\includegraphics[width=0.185\linewidth]{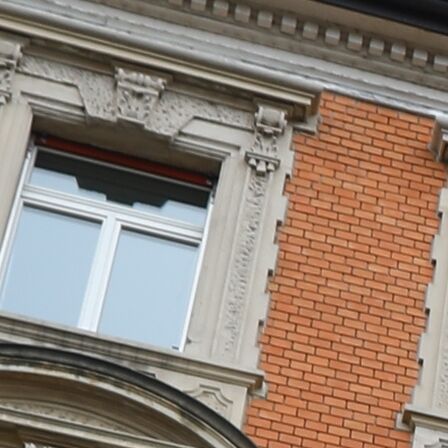}} \\ \vspace{-.3cm}

\subfloat[RAW]{
\includegraphics[width=0.185\linewidth]{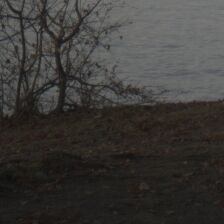}}
\hfill
\subfloat[AWNet]{
\includegraphics[width=0.185\linewidth]{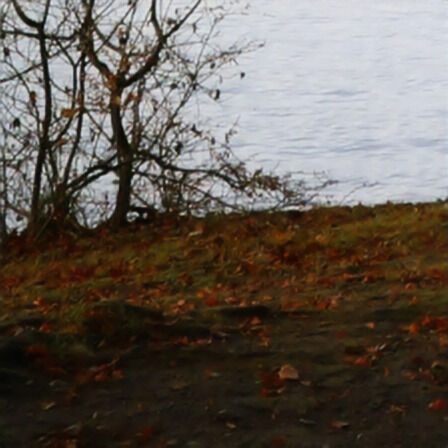}}
\hfill
\subfloat[PyNet]{
\includegraphics[width=0.185\linewidth]{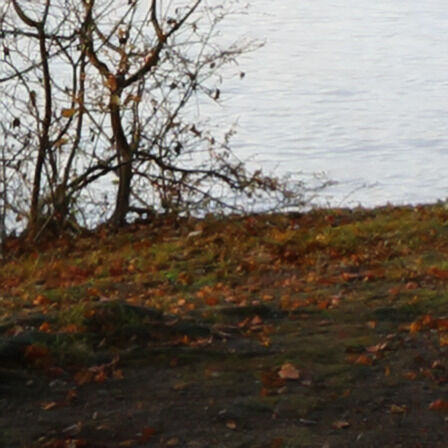}}
\hfill
\subfloat[Ours]{
\includegraphics[width=0.185\linewidth]{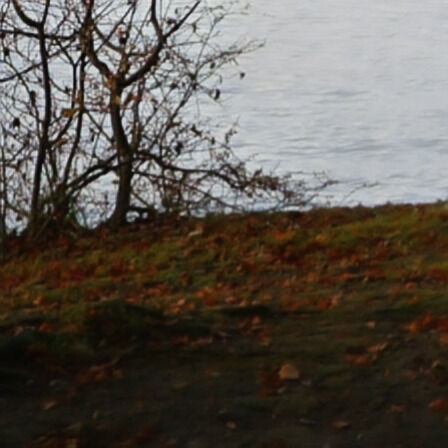}}
\hfill
\subfloat[DSLR (target)]{
\includegraphics[width=0.185\linewidth]{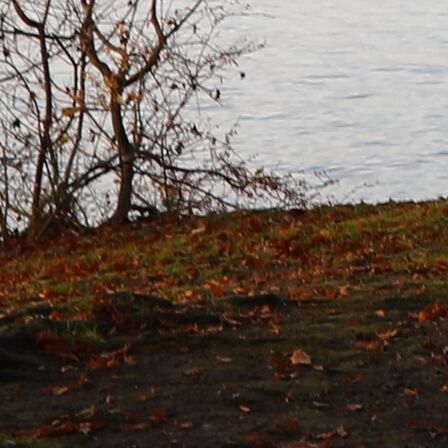}} 

\end{center}

   \caption{Comparison image details}
\label{fig:detail}
\end{figure}

\subsection{Drawbacks of this Method}
While FLOPs and memory usage (RAM / VRAM) are low, parameter count is increased and
therefore model-size and HDD space requirements are increased.

For large batch-sizes and small images, grouped convolutions (used in the HyperConvolution) 
can be less efficient than standard convolutions (because of potentially limited parallelism).

\section{Experiments}
In this section we present our experimental findings on the Zurich RAW to DSLR task and the SIDD task.
Firstly, we observe the performance of our networks as a function of the scaled size of fully-connected
layers in a HyperNetwork, in Sec. \ref{subsec:dd}. Secondly, we build a large network making use of the proposed HyperConvolutions and compare to state of the art 
in Sec. \ref{subsec:large}. Finally, we substitute convolution layers in a VDN \cite{yue2019variational} for our HyperConvolutions and show matched fidelity
at reduced computational cost in Sec. \ref{subsec:SIDD}.

All experiments were run using PyTorch \cite{pytorch}. Training details are given in the appendix, along with code used for the key functionalities described.

\subsection{Data-Sets}
\paragraph{Zurich Raw to DSLR}
We perform our experiments on the Zurich Raw to DSLR data-set \cite{ignatov2020replacing} (ZRR). This data-set consists of 48'043 image pairs 
of resolution 448. 
Each pair contains one RAW image taken with a mobile phone camera (Huawei P20 Pro) and one fully post-processed
JPEG output of a high-end DSLR camera (Canon 5D Mark IV). The images in such a pair cover the same scene under the 
same view-point and can be assumed to be well-aligned. 
We follow \cite{ignatov2020replacing} and transform RAW images from shape $(1,448,448)$ to $(4, 224, 224)$ by transposing the Bayer pattern
into the channel dimension.

The task is the prediction of the DSLR output from the mobile phone camera RAW. In doing this, the algorithm 
must perform the function of the full ISP pipeline, including demoisaicing, denoising, contrast adaptation 
and white-balancing.

We choose this data-set, because it includes these various aspects in one place and because it decouples the 
processing pipeline from the network architecture design, which is our primary focus.  
Furthermore the authors of \cite{ignatov2020replacing} compare a number of different baseline architectures.

This data-set was featured in the ECCV 2020 AIM Learned Smartphone ISP Challenge \cite{ignatov2020aim}. At the time of writing we could find sufficient detail on \cite{dai2020awnet} and \cite{kim2020pynet} publicly available for a comparison of single network fidelity and computational cost.

\paragraph{Smartphone Image Denoising Dataset}
SIDD-Medium \cite{abdelhamed2018high} is a widely used image denoising dataset consisting of 320 image pairs of noisy and ground-truth images collected from an array of 5 smartphone cameras in various illumination settings. 
There are two variants, one using RAW input the other sRBG input, in both the aim is to predict a denoised sRGB output. To complement our experiments on ZRR, we use the sRBG to sRGB variant.

\subsection{Double-Descent Generalization}
\label{subsec:dd}
In this series of experiments we take networks containing HyperConvolutions 
and vary the number of hidden units in the MLP inside the HyperConvolutions.
Furthermore we take standard ConvNets with the same architecture and
vary the number of channels in hidden layers.

We are interested in two things:
1) As a function of the number of parameters, how does the generalization error
change?
2) As a function of the number of parameters, how does the number of FLOPs
the network requires change?

Our goal is to find out, whether it is possible to decrease generalization error
without needing to increase FLOPs. This is motivated by the observation that
many networks generalize better, when they have more parameters (see Sec.\ref{sec:dd})
and the realization that HyperConvolutions can break the fixed parameter
to FLOP ratio of standard convolutions (see Sec. \ref{sec:method})

\subsubsection{Network Layout}
\begin{figure*}[th]
\begin{center}
\includegraphics[width=0.75\linewidth]{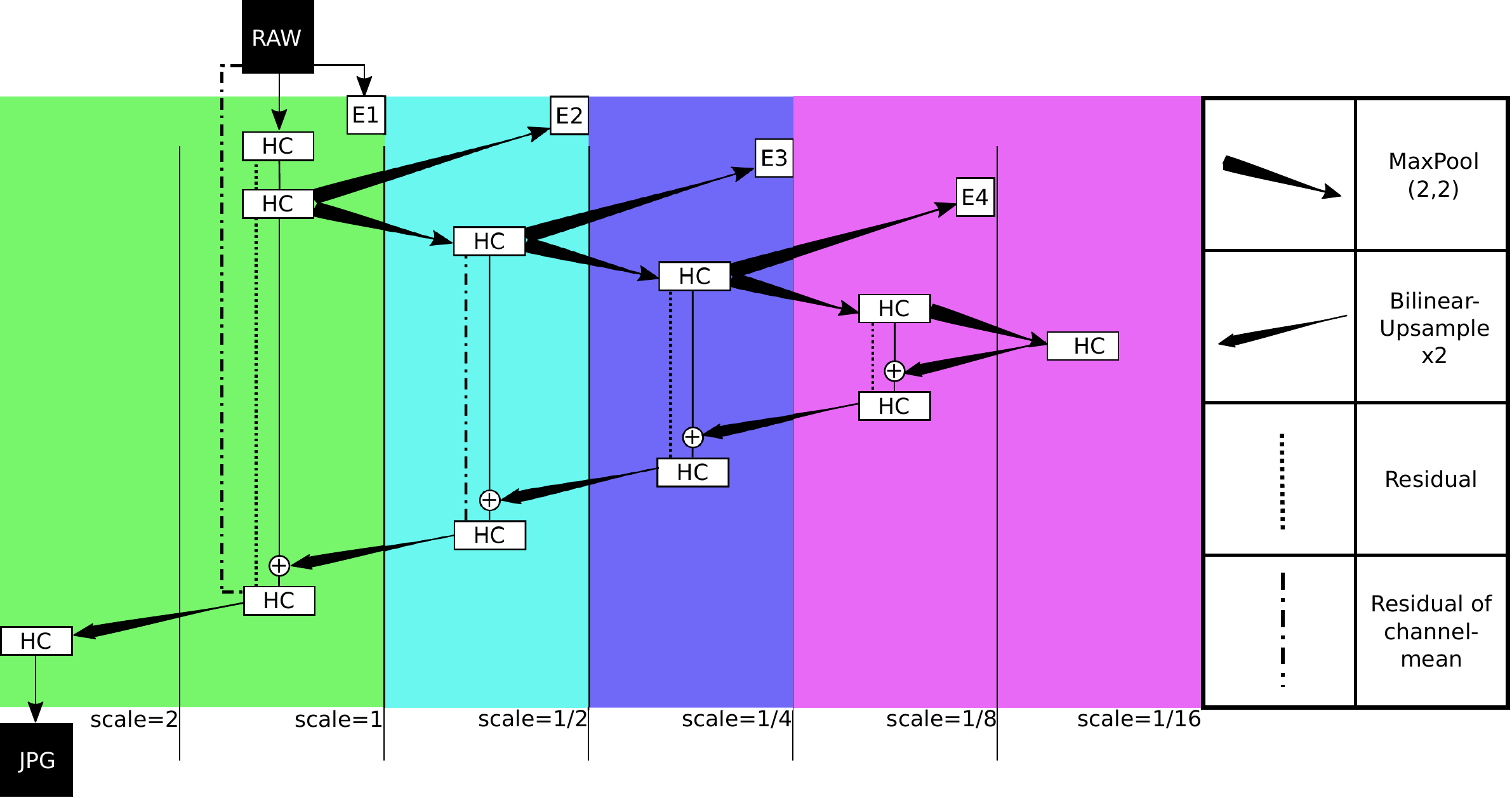}

\end{center}
   \caption{Layout of the UNet-like architecture. The boxes labeled `E' are placeholders for embedding networks.
   The colored backdrop indicates which embedding network E gives the filter input to a given HyperConvolution HC.}
\label{fig:unet}
\end{figure*}

\begin{figure}[ht]
\begin{center}

   \includegraphics[width=0.8\linewidth]{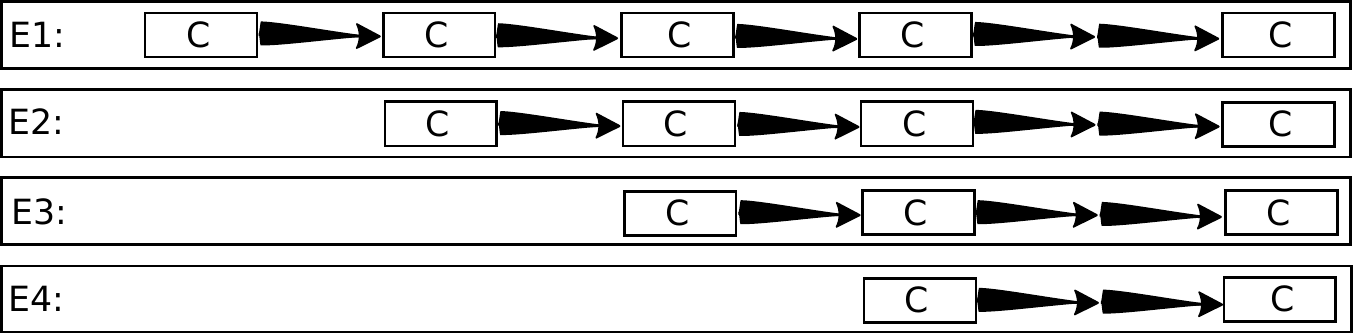}
   \end{center}
   \caption{Layouts of embedding networks. `C' stands for $3\times3$ convolution layers. }
\label{fig:embeddings}
\end{figure}

We run this experiment for a UNet-like architecture \cite{ronneberger2015u}.

In this UNet, 
each occurrence of a convolutional
layer is replaced with a HyperConvolution. The number of feature maps in these layers is proportional to $n_\text{fwd}$. 
The filter input to this HyperConvolution is computed by embedding networks (various ones at different scales). 
The number 
of feature maps in the layers of all our embedding networks start at $n_\text{embed}$ and double 
at each subsequent layer. Finally $n_\text{hid}$ indicates the number of units in the two hidden layers
of the MLP that predicts the filters in each HyperConvolution.
For an illustration see Fig.~\ref{fig:unet}, for code see the appendix.  

All convolutions and HyperConvolutions use filter size $3\times 3$ and the GELU \cite{hendrycks2016gaussian} non-linearity,
except for the first two HyperConvolution layers which use a hard-sigmoid. We set $n_\text{fwd}=8$ and  $n_\text{embed}=8$.
Further details are given in the appendix.

Note that whenever we report FLOPs and parameter numbers we count both the forward network,
as well as the filter prediction network.

\subsubsection{Results}
Figs.~\ref{fig:mainresult},~\ref{fig:mainresultb} show as a function of the number of parameters the test error (1-MS-SSIM) on ZRR
and the FLOPs of the considered network and the same network without HyperConvolution.

The proposed HyperConvolution exhibits a
generalization curve consistent with double-descent generalization
as we increase number of units in the MLP, at negligible FLOP-count increase. 
Further we see that with
standard convolutions the FLOP-count increases steeply with decreasing generalization error, and that at the same
FLOP count the over-parameterized HyperConvolution network performs much better. 

A plausible underlying explanation for these observations is that the number of parameters 
(the amount of information the network can store about the training data) is the limiting factor
in the learning process. 
This motivates the development of mechanisms
to increase parameter-density in ConvNets like the one given in this paper.
 
\subsection{Large Network}
\label{subsec:large}
In this experiment we build a larger network of the same layout as given
in Fig.~\ref{fig:unet} and compare its performance to state-of-the-art architectures trained on the same input.

\subsubsection{Results}
In Tab.~\ref{tab:bigtab} we list performance metrics of several network
sizes alongside current state-of-the-art. The various networks listed 
as `Ours' are variants with different settings of $n_\text{fwd}$, 
$n_\text{embed}$ and $n_\text{hid}$. 

CPU time was measured by running a single full-scale image through the evaluation 
code (where available) on a Ryzen 3700x CPU. The given memory usage is the maximal memory use 
the pytorch-profiler reports for a convolution. We counted FLOPs using
the pytorch-flops-counter\footnote{\href{https://github.com/sovrasov/flops-counter.pytorch}{https://github.com/sovrasov/flops-counter.pytorch}}.

Two observations are of particular note. Firstly, our largest network 
exceeds SSIM and MS-SSIM of much more costly networks (in terms of compute time, FLOP count, and
memory use). Secondly, the `forward' network used here is comparatively small. 

The second observation raises the question, whether 
there is a limit beyond which decreasing $n_\text{fwd}$  (the width of the forward network) 
cannot be compensated for, by increasing $n_\text{hid}$ (the size of the MLPs in the HyperConvolutions). 
To answer this question we run a further experiment with $n_\text{fwd}=8$
and $n_\text{hid} =  4000$ (all else the same). We find that this network under-performs compared to
the wider networks with fewer parameters (see Tab.~\ref{tab:bigtab}).

This
indicates that it is not sufficient to have many parameters. We speculate
that in a network with very low $n_\text{fwd}$, the narrow forward
layers form an information bottleneck \cite{shwartz2017opening}, that prevents sufficient
information about the input from propagating through the network to the
output (eventhough there would be sufficiently many parameters to `interpolate'
the input data-set).

In Figs.~\ref{fig:detail},~\ref{fig:pics} we show sample outputs of our network, next to 
the outputs of prior approaches. 
 
\subsection{Efficient Image Denoising}

\label{subsec:SIDD}
To verify a broader applicability of our proposed method, in this section we follow closely the setup (architecture, loss, training) of a well established method (VDN \cite{yue2019variational}), but we replace convolution layers with the HyperConvolutions proposed in Sec.~\ref{sec:method}. The VDN architecture is a variational method where the network predicts from a noisy input both an estimate of the noise-free image and the noise variance per pixel. We test whether a narrower network (fewer channels per layer) with HyperConvolutions can match the fidelity of the standard VDN at reduced computational cost. 

The setup is identical to the VDN reference implementation on github\footnote{\href{https://github.com/zsyOAOA/VDNet}{https://github.com/zsyOAOA/VDNet}} (with the exception of a bug-fix relating to learning-rate decay). Each convolution with $c$ channels is replaced with a HyperConvolution layer with $\mathtt{int}(c/3)$ channels (and $\mathtt{int}(c/2)$ channels in a second experiment). The HyperNetwork that predicts the filter of any given HyperConvolution receives the same input as the HyperConvolution itself and consists of a 3-layer CNN, each layer of which has ReLU non-linearity, stride 2, and their respective channel numbers are 32, 64 and 128 (for code consider the appendix). 

The results can be found in Table~\ref{tab:sidd}. The FLOP count of our modified VDN is ca. $6.8\times$ lower, memory use and CPU time are also reduced. We indicate memory use and CPU time for a 1 MPix Srgb image as is customary for this benchmark and again use a Ryzen 3700x CPU.  The validation set results were performed using the VDN reference code, test results come from the SIDD benchmark website (hence difference in SSIM). 

In summary, our modified network matches the fidelity of the standard VDN (it is on-par in PSNR and slightly better in SSIM as for the previous task) at substantially reduced computational cost.

\section{Discussion}
Often machine learning models are evaluated with the perspective that 
fewer parameters are inherently better. This makes sense given the `classical' bias-variance generalization
curve, because in this settings small models with good predictive ability are
the ones with the best inductive biases; their structure best models the data.

In the days of interpolating machine learning models, for which
parameter abundance (or high dimensionality) is itself a useful inductive bias, 
it is arguably time to shift our focus to other metrics of complexity, such as memory use and run-time.

A key result of this paper 
is that more accurate models with more parameters need not have higher memory requirements
and computational cost.

\section{Conclusions}
In this paper we show that additional parameters can improve generalization
in ConvNets with minor impact on the compute cost of the network. The central idea
is to use HyperNetworks containing fully-connected layers to predict convolutional filters
from input images. The FLOP-count of such fully-connected layers does not scale with
input image size. This allows increasing parameter count independently from FLOP count 
in the large image limit. Using this insight we achieve state-of-the-art single-network SSIM and MS-SSIM 
on the ZRR task with fewer FLOPs and
smaller memory footprint than the best previous architectures \cite{ignatov2020replacing, dai2020awnet}. 

To verify the broader applicability of our proposed method, we further demonstrate
that its use in the existing network `VDN' \cite{yue2019variational} can substantially reduce computational cost
while matching fidelity in the SIDD task.

For the tasks of
denoising and enhancing RAW photos shot with small mobile camera sensors, FLOP-count is crucial for
practical utility, as the mobile setting demands conservative energy and time use. 
Energy-efficiency is also of high relevance when the environmental impact
of training deep learning models is considered. 
Our approach adds a new tool to the repertoire of energy-efficient image enhancement,
as well as offering a new avenue of attack on other parameter-count limited tasks.

{\small
\bibliographystyle{ieee_fullname}
\bibliography{biblio}
}

\clearpage

\section*{Appendix: HyperNetworks with Double-Descent Generalization at Fixed FLOP-Count \\ Enable Fast Neural Image Enhancement}

\subsection*{Training Details}
\subsubsection*{Double Descent Plots}
In table \ref{tab:training1} we list the training hyperparameters.  Where multiple values were considered, they are given in 
braces, with the preferred value in bold. 
For hyperparameter selection we trained and validated on a subset of the training data
and chose the best values. 

The loss-functions we use are mean squared error (MSE) and multi-scale structural similarity (MS-SSIM) \cite{wang2003multiscale}. 
\begin{table}[h]
\centering
\begin{tabular}{|l|r|}
\hline
Hyperparameter & Value \\
\hline
\hline
optimizer & ADAM \cite{kingma2014adam} \\
batch size & 8 \\
learning rate & $\lbrace 2e\text{-}6, \bold{5e\text{-}6}, 1e\text{-}5 \rbrace$ \\
$\beta_1$ & $\lbrace 0.5, \bold{0.7}, 0.9 \rbrace$ \\
$\beta_2$ & $\lbrace 0.9, \bold{0.95}, 0.99 \rbrace$ \\
epochs & 30 \\
$n_\text{fwd}$ & 8 \\
$n_\text{embed}$ & 8 \\
$n_\text{hid}$ & varies \\
loss & MSE + (1-MS-SSIM) \\
\hline
\end{tabular}
\caption{Hyperparameters for the double descent experiments. Bold values were used, values in braces were considered for selection.
For further details see flow-text.}
\label{tab:training1}
\end{table}
\subsubsection*{Large Network}
In Tab.~\ref{tab:training1} we list the training hyperparameters. At epoch 25 we 
switch from ADAM \cite{kingma2014adam} to SGD.
From epoch 20 onward, we add a VGG-loss at every second batch based on VGG-19 \cite{simonyan2014very}.
\begin{table}[h]
\centering
\begin{tabular}{|l|r|}
\hline
Hyperparameter & Value  \\
\hline
\hline
optimizer & ADAM \cite{kingma2014adam} and SGD \\
batch size & 4 \\

epochs & 30 \\
$n_\text{fwd}$ & $\lbrace 16, 32, \bold{64}\rbrace$ \\
$n_\text{embed}$ & $\lbrace 16, \bold{32}\rbrace$ \\
$n_\text{hid}$ & 2048 \\
loss & MSE + (1 - MS-SSIM) (+ VGG)\\
\hline
\end{tabular}

\caption{Hyperparameters for the large network (where different from double descent experiments). 
Bold values were used, values in braces were considered for selection.
For further details see flow-text.}
\end{table}

\newpage

\subsection*{Code}
For convenience we provide some key code snippets below. Please consider the license before use. 
\begin{lstlisting}[basicstyle=\ttfamily\tiny]
Copyright 2021 Lorenz K. Muller, Huawei Technologies

Licensed under the Apache License, Version 2.0 (the "License");
you may not use this file except in compliance with the License.
You may obtain a copy of the License at

    http://www.apache.org/licenses/LICENSE-2.0

Unless required by applicable law or agreed to in writing, software
distributed under the License is distributed on an "AS IS" BASIS,
WITHOUT WARRANTIES OR CONDITIONS OF ANY KIND, either express or implied.
See the License for the specific language governing permissions and
limitations under the License.
\end{lstlisting}

\begin{table*}
\begin{lstlisting}[language=Python, basicstyle=\ttfamily\tiny]
class HyperConv(nn.Module):
    def __init__(self, n_in_forward, n_in_hyper, n_out, f_size=3, bias=False, gain=False, n_hid=256, decompose=0,
                 op_count=False):
        '''
        A hyper-network convolutional filter. We get two sets of input feature maps. The first set we average pool
        over the whole image to linearly predict filters. These filters we use to convolve the second set.
        :param n_in_forward: number of channels of the input to be fed forward
        :param n_in_hyper: number of channels of the input to the hyper-network layer
        :param n_out: number of learned filters / number of output channels
        :param f_size: size of the learned filter
        :param bias: flag to add bias
        :param gain: flag to add multiplicative bias
        :param n_hid: size of the hidden layers in the MLP
        :param decompose: if >0 the filters are predicted as a low-rank decomposition (not used in paper)
        :param op_count: flag to enable a variant that is friendly to the op-counting script we used
        '''
        super(HyperConv, self).__init__()
        self.n_in_forward = n_in_forward
        self.n_in_hyper = n_in_hyper
        self.n_out = n_out
        self.f_size = f_size
        self.pool = nn.AdaptiveMaxPool2d(1)
        self.pad = nn.ReflectionPad2d(f_size // 2)
        self.n_hid = n_hid
        self.decompose = decompose
        self.op_count = op_count
        if decompose < 1:
            self.outf = nn.Sequential(nn.Linear(n_in_hyper, n_hid),
                                      nn.ReLU(),
                                      nn.Linear(n_hid, n_hid),
                                      nn.ReLU(),
                                      nn.Linear(n_hid, f_size * f_size * n_in_forward * n_out))
        else:
            self.outf1 = nn.Sequential(nn.Linear(n_in_hyper, n_hid),
                                      nn.ReLU(),
                                      nn.Linear(n_hid, n_hid),
                                      nn.ReLU(),
                                      nn.Linear(n_hid, f_size * f_size * 1 * n_out * decompose))
            self.outf2 = nn.Sequential(nn.Linear(n_in_hyper, n_hid),
                                      nn.ReLU(),
                                      nn.Linear(n_hid, n_hid),
                                      nn.ReLU(),
                                      nn.Linear(n_hid, f_size * f_size * n_in_forward * 1 * decompose))

        if op_count:
            self.fwd_conv = nn.Conv2d(self.n_in_forward,self.n_out,
                                      kernel_size=f_size, groups=1)

        self.bias = None if not bias else nn.Sequential(nn.Linear(n_in_hyper, n_out))
        self.gain = None if not gain else nn.Sequential(nn.Linear(n_in_hyper, n_out))

        self.const_f = nn.Parameter(data=torch.randn([1, n_out, n_in_forward, f_size, f_size])/np.sqrt(self.n_in_forward*f_size*f_size/2)/2, requires_grad=True)
        self.norm = np.sqrt(self.n_in_forward * np.pi / f_size / f_size)

    def forward(self, x):
        '''
        :param x: a list of the form [x_hyper, x_forward]
        :return: the convolved x_forward
        '''
        xfilter, xforward = x
        # predict a set of conv filters for each sample
        f = self.pool(xfilter)  # pool over the full image (becomes independent of input image size)
        if self.bias is not None:
            b = self.bias(f.view(-1, self.n_in_hyper))
            b = torch.clamp(b, -.1, .1)
        if self.gain is not None:
            g = self.gain(f.view(-1, self.n_in_hyper))
            g = torch.clamp(g, -.9, .1) + 1.

        if not self.decompose:
            f = self.outf(f.view(-1, self.n_in_hyper))  # flatten pooled image and pass through linear layer
        else:
            f1 = self.outf1(f.view(-1, self.n_in_hyper)).view(-1, self.n_out, 1,
                                                              self.f_size, self.f_size, self.decompose)
            f2 = self.outf2(f.view(-1, self.n_in_hyper)).view(-1, 1, self.n_in_forward,
                                                              self.f_size, self.f_size, self.decompose)
            f = (f1*f2).mean(5)

        f = f / self.norm
        f = f.view(-1, self.n_out, self.n_in_forward, self.f_size, self.f_size).contiguous()
        f = f / (f.abs().sum([3, 4], keepdims=True) + 1e-12)/2 + self.const_f

        # we reflection pad the forward input (to avoid odd looking image boundaries)
        xforward = self.pad(xforward)

        # here we need to split the batch up, to apply an individual conv-filter to each sample
        # we do this using the 'groups' keyword of the convolution with the appropriate .view
        # in essence this does F.conv2d(xforward, predicted_filter) but the filter varies per sample
        xf = F.conv2d(
            xforward.view(1, xforward.size()[0] * self.n_in_forward, xforward.size(2), xforward.size(3)),
            f.view(xforward.size()[0] * self.n_out, self.n_in_forward, self.f_size, self.f_size),
            groups=xforward.size()[0])
        if self.op_count:
            xf = self.fwd_conv(xforward)
        xf = xf.view(xforward.size()[0], self.n_out, xf.size(2), xf.size(3))
        if self.bias is not None:
            xf = xf + b.view(xforward.size()[0], self.n_out, 1, 1)
        if self.gain is not None:
            xf = xf * g.view(xforward.size()[0], self.n_out, 1, 1)

        return xf
\end{lstlisting}
\caption{Code for Hyperconvolution Layer.}
\end{table*}

\begin{table*}
\begin{lstlisting}[language=Python, basicstyle=\ttfamily\tiny]
class HyperUNet(nn.Module):
    def __init__(self, n_input_channels, n_fwd=32, embed_base=16, n_out=3, n_hid=256, op_count=False):
        super(HyperUNet, self).__init__()
        self.n_input_channels = n_input_channels
        self.n_fwd = n_fwd
        self.embed_base = embed_base

        # embedding at input scale
        self.embed = nn.Sequential(nn.Conv2d(n_input_channels, embed_base, 3, padding=1),  # 224
                                   nn.MaxPool2d(2, 2), nn.GELU(),
                                   nn.Conv2d(embed_base, embed_base * 2, 3, padding=1),  # 112
                                   nn.MaxPool2d(2, 2), nn.GELU(),
                                   nn.Conv2d(embed_base * 2, embed_base * 4, 3, padding=1),  # 56
                                   nn.MaxPool2d(2, 2), nn.GELU(),
                                   nn.Conv2d(embed_base * 4, embed_base * 8, 3, padding=1),  # 28
                                   nn.MaxPool2d(4, 4), nn.GELU(),
                                   nn.Conv2d(embed_base * 8, embed_base * 16, 3, padding=1))  # 7
        # embedding at input scale // 2
        self.embed2 = nn.Sequential(nn.Conv2d(n_fwd, embed_base, 3, padding=1, stride=2),  # 112
                                    nn.MaxPool2d(2, 2), nn.GELU(),
                                    nn.Conv2d(embed_base, embed_base * 2, 3, padding=1),  # 56
                                    nn.MaxPool2d(2, 2), nn.GELU(),
                                    nn.Conv2d(embed_base * 2, embed_base * 4, 3, padding=1),  # 28
                                    nn.MaxPool2d(4, 4), nn.GELU(),
                                    nn.Conv2d(embed_base * 4, embed_base * 8, 3, padding=1))  # 7
        # embedding at input scale // 4
        self.embed4 = nn.Sequential(nn.Conv2d(n_fwd//2, embed_base//2, 3, padding=1),  # 56
                                    nn.MaxPool2d(2, 2), nn.GELU(),
                                    nn.Conv2d(embed_base//2, embed_base * 2, 3, padding=1),  # 28
                                    nn.MaxPool2d(4, 4), nn.GELU(),
                                    nn.Conv2d(embed_base * 2, embed_base * 4, 3, padding=1))  # 7
        # embedding at input scale // 8
        self.embed8 = nn.Sequential(nn.Conv2d(n_fwd//2, embed_base//2, 3, padding=1),  # 56
                                    nn.MaxPool2d(4, 4), nn.GELU(),
                                    nn.Conv2d(embed_base//2, embed_base * 2, 3, padding=1))  # 14

        # hyper-convs at input scale
        self.hyper_conv0 = HyperConv(n_input_channels, embed_base * 16, n_fwd, f_size=3, bias=True, n_hid=n_hid, op_count=op_count)
        self.hyper_conv1 = HyperConv(n_fwd, embed_base * 16, n_fwd, bias=True, f_size=3, n_hid=n_hid, op_count=op_count)
        self.hyper_conv2 = HyperConv(n_fwd, embed_base * 16, n_fwd, bias=True, f_size=3, n_hid=n_hid, op_count=op_count)
        # hyper-convs at input scale *2
        self.hyper_conv4 = HyperConv(n_fwd, embed_base * 16, n_out, bias=True, f_size=3, n_hid=n_hid, op_count=op_count)
        # hyper-convs at input scale //2
        self.hyper_conv1_2 = HyperConv(n_fwd, embed_base * 8, n_fwd//2, bias=True, f_size=3, n_hid=n_hid//2, op_count=op_count)
        self.hyper_conv2_2 = HyperConv(n_fwd//2, embed_base * 8, n_fwd, bias=True, f_size=3, n_hid=n_hid//2, op_count=op_count)        
        # hyper-convs at input scale //4
        self.hyper_conv1_4 = HyperConv(n_fwd//2, embed_base * 4, n_fwd//2, bias=True, f_size=3, n_hid=n_hid//2, op_count=op_count)
        self.hyper_conv2_4 = HyperConv(n_fwd//2, embed_base * 4, n_fwd//2, bias=True, f_size=3, n_hid=n_hid//2, op_count=op_count)
        # hyper-convs at input scale //8
        self.hyper_conv1_8 = HyperConv(n_fwd//2, embed_base * 2, n_fwd//2, bias=True, f_size=3, n_hid=n_hid//2, op_count=op_count)
        self.hyper_conv2_8 = HyperConv(n_fwd//2, embed_base * 2, n_fwd//2, bias=True, f_size=3, n_hid=n_hid//2, op_count=op_count)
        # hyper-convs at input scale //16
        self.hyper_conv1_16 = HyperConv(n_fwd//2, embed_base * 2, n_fwd//2, bias=True, f_size=3, n_hid=n_hid//2, op_count=op_count)

        # scaling operations
        self.pool = nn.MaxPool2d(2, 2)
        self.scale = nn.Upsample(scale_factor=2.0, mode='bilinear', align_corners=True)


    def forward(self, x):
        # embed image
        e = self.embed(x)

        # scale * 1
        y = self.hyper_conv0([e, x])
        y0 = torch.clamp(y,0,1)
        y = self.hyper_conv1([e, y0]) + y0
        y0 = torch.clamp(y,0,1)
        # scale // 2
        y2 = self.pool(y0)
        e2 = self.embed2(y0)
        y2 = F.gelu(self.hyper_conv1_2([e2, y2]))
        # scale // 4
        y4 = self.pool(y2)
        e4 = self.embed4(y4)
        y4 = F.gelu(self.hyper_conv1_4([e4, y4]))
        # scale // 8
        y8 = self.pool(y4)
        e8 = self.embed8(y8)
        y8 = F.gelu(self.hyper_conv1_8([e8, y8]))
        # scale // 16
        y16 = self.pool(y8)
        e16 = e8
        y16 = F.gelu(self.hyper_conv1_16([e16, y16]))
        # scale // 8
        y8 = F.gelu(self.hyper_conv2_8([e8, y8 + self.scale(y16)*.25]) + y8)
        # scale // 4
        y4 = F.gelu(self.hyper_conv2_4([e4, y4 + self.scale(y8)*.25]) + y4)
        # scale // 2
        y2 = F.gelu(self.hyper_conv2_2([e2, y2 + self.scale(y4)*.25]) + y2.mean(1).unsqueeze(1))
        # scale * 1
        y = F.gelu(self.hyper_conv2([e, y0 + self.scale(y2)*.25]) + y0)
        y = y + x.mean(1).unsqueeze(1)
        # scale * 2
        y = self.scale(y)
        y = self.hyper_conv4([e, y])
        return torch.sigmoid(y)

\end{lstlisting}
\caption{Code for UNet-like architecture.}
\end{table*}

\begin{table*}
\begin{lstlisting}[language=Python, basicstyle=\ttfamily\tiny]
class HyperConvEmbed(nn.Module):
    def __init__(self, n_in_forward, n_out, f_size=3, bias=False, gain=False, n_hid=64):
        '''
        A hyper-network convolutional filter. First we 'embed' the input and average pool
        over the whole image predict filters. These filters we use to convolve the original input.
        :param n_in_forward: number of channels of the input to be fed forward
        :param n_out: number of learned filters / number of output channels
        :param f_size: size of the learned filter
        :param bias: flag to add bias
        :param gain: flag to add multiplicative bias
        :param n_hid: size of the hidden layers in the MLP
        '''
        super(HyperConvEmbed, self).__init__()
        self.n_in_forward = n_in_forward
        self.n_out = n_out
        self.f_size = f_size
        self.pool = nn.AdaptiveAvgPool2d(1)
        self.pad = nn.ReflectionPad2d(f_size // 2)
        self.n_hid = n_hid

        embed_base = 32
        self.n_in_hyper = n_in_hyper = embed_base * 4
        self.embed = nn.Sequential(nn.Conv2d(self.n_in_forward, embed_base, 3, padding=1, stride=2),
                                   nn.ReLU(),
                                   nn.Conv2d(embed_base, embed_base * 2, 3, padding=1, stride=2),
                                   nn.ReLU(),
                                   nn.Conv2d(embed_base * 2, embed_base * 4, 3, padding=1, stride=2))

        self.outf = nn.Sequential(nn.Linear(n_in_hyper, n_hid),
                                  nn.LeakyReLU(),
                                  nn.Linear(n_hid, n_hid),
                                  nn.LeakyReLU(),
                                  nn.Linear(n_hid, f_size * f_size * n_in_forward * n_out))

        self.bias = None if not bias else nn.Sequential(nn.Linear(n_in_hyper, n_out))

        self.gain = None if not gain else nn.Sequential(nn.Linear(n_in_hyper, n_out))

        self.const_f = nn.Parameter(data=torch.randn([1, n_out, n_in_forward, f_size, f_size]) / np.sqrt(
            self.n_in_forward * f_size * f_size / 2) / 2, requires_grad=True)
        self.norm = np.sqrt(self.n_in_forward * np.pi / f_size / f_size)

    def forward(self, x):
        '''
        :param x: forward input
        :return: the convolved x
        '''
        xfilter = self.embed(x)
        xforward = x
        # predict a set of conv filters for each sample
        f = self.pool(xfilter)  # pool over the full image (becomes independent of input image size)
        if self.bias is not None:
            b = self.bias(f.view(-1, self.n_in_hyper))
            b = torch.clamp(b, -.1, .1)
        if self.gain is not None:
            g = self.gain(f.view(-1, self.n_in_hyper))
            g = torch.clamp(g, -.9, .1) + 1.

        f = self.outf(f.view(-1, self.n_in_hyper))  # flatten pooled image and pass through linear layer

        f = f / self.norm
        if xforward.size()[0] != 1:
            f = f.view(-1, self.n_out, self.n_in_forward, self.f_size, self.f_size)
            f = f / (f.abs().sum((3, 4), keepdim=True) + 1e-12) / 2 + self.const_f
        else:
            f = f.view(self.n_out, self.n_in_forward, self.f_size, self.f_size)
            f = f / (f.abs().sum((2, 3), keepdim=True) + 1e-12) / 2 + self.const_f.squeeze(0)

        # we reflection pad the forward input (to avoid odd looking image boundaries)
        xforward = self.pad(xforward)

        # here we need to split the batch up, to apply an individual conv-filter to each sample
        # we do this using the 'groups' keyword of the convolution with the appropriate .view
        # in essence this does F.conv2d(xforward, predicted_filter) but the filter varies per sample
        if xforward.size()[0] != 1:
            xf = F.conv2d(
                xforward.view(1, xforward.size()[0] * self.n_in_forward, xforward.size(2), xforward.size(3)),
                f.view(xforward.size()[0] * self.n_out, self.n_in_forward, self.f_size, self.f_size),
                groups=xforward.size()[0])
        else:
            xf = F.conv2d(xforward, f)

        xf = xf.view(xforward.size()[0], self.n_out, xf.size(2), xf.size(3))
        if self.bias is not None:
           xf = xf + b.view(xforward.size()[0], self.n_out, 1, 1)
        if self.gain is not None:
           xf = xf * g.view(xforward.size()[0], self.n_out, 1, 1)

        return xf
        
\end{lstlisting}
\captionsetup{justification=centering}
\caption{Code for `drop-in' HyperConvolution layer used in VDN (includes embedding) assuming `same' padding mode. }
\end{table*}

\newpage
\subsection*{Large-scale Image outputs}
Of our method (compressed due to file-size limitations)

\begin{landscape}
\begin{figure}
\centering
\includegraphics[height=0.95\textheight]{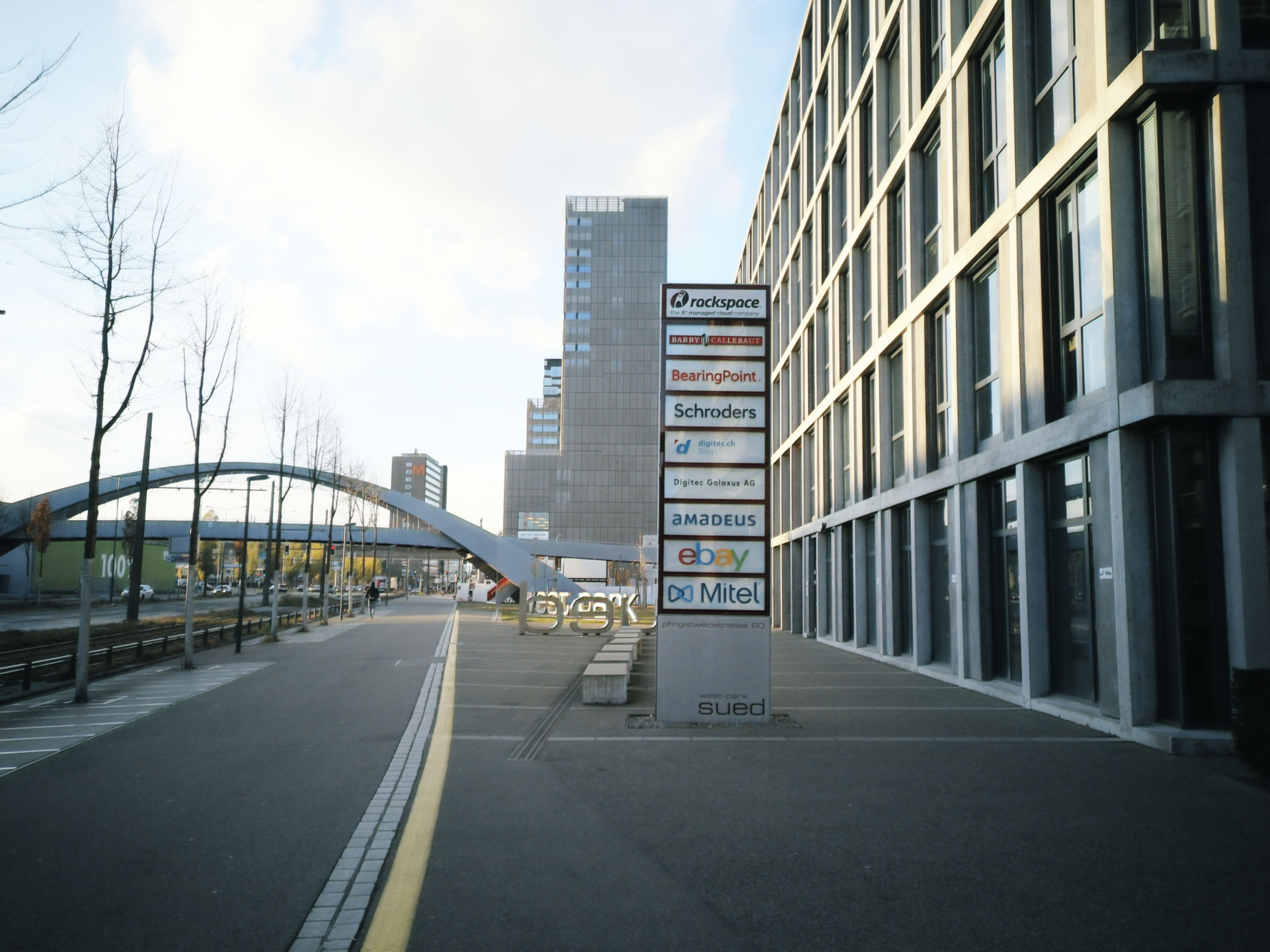}
\end{figure}
\end{landscape}

\begin{landscape}
\begin{figure}
\centering
\includegraphics[height=0.95\textheight]{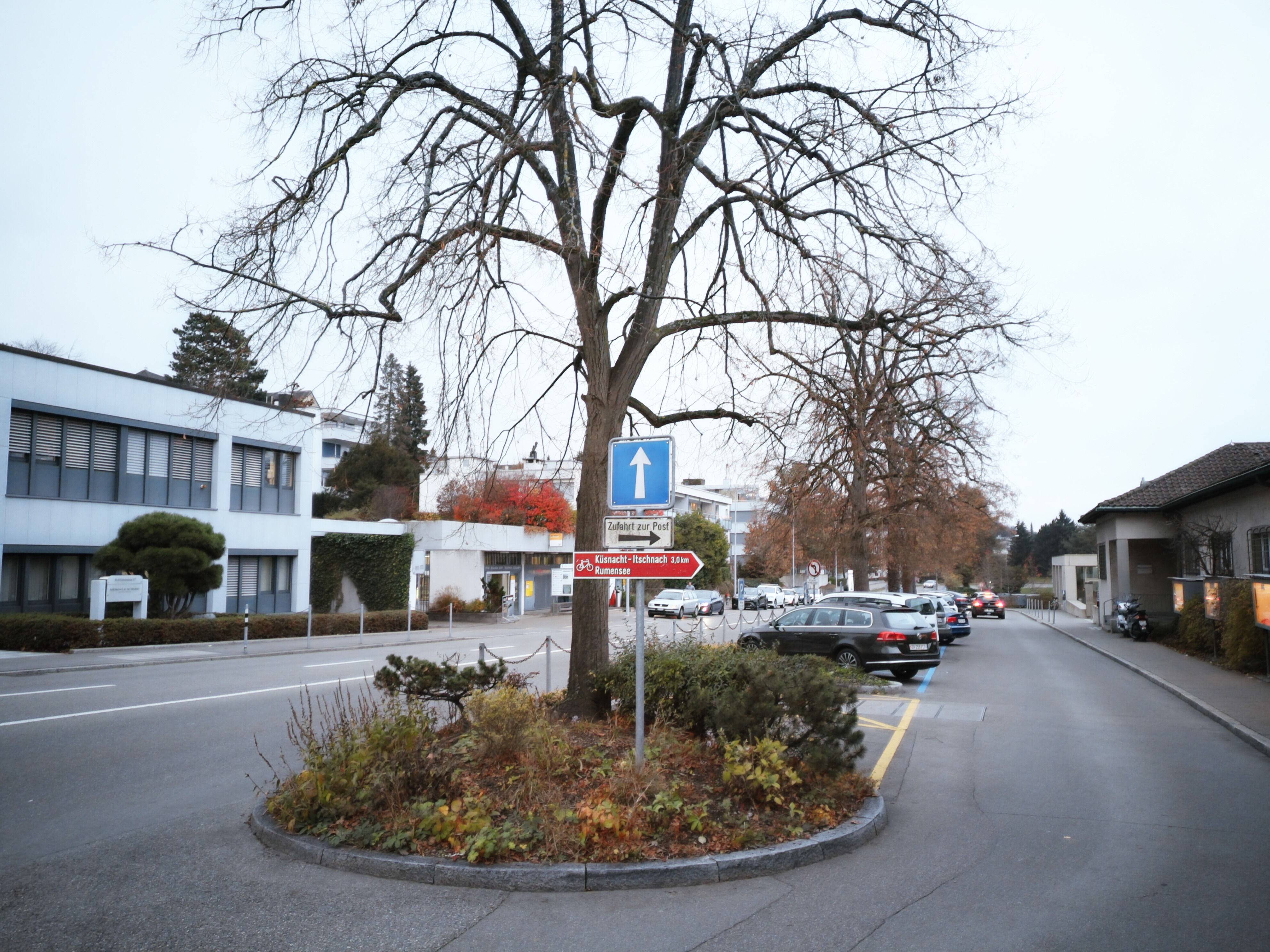}
\end{figure}
\end{landscape}

\begin{landscape}
\begin{figure}
\centering
\includegraphics[height=0.95\textheight]{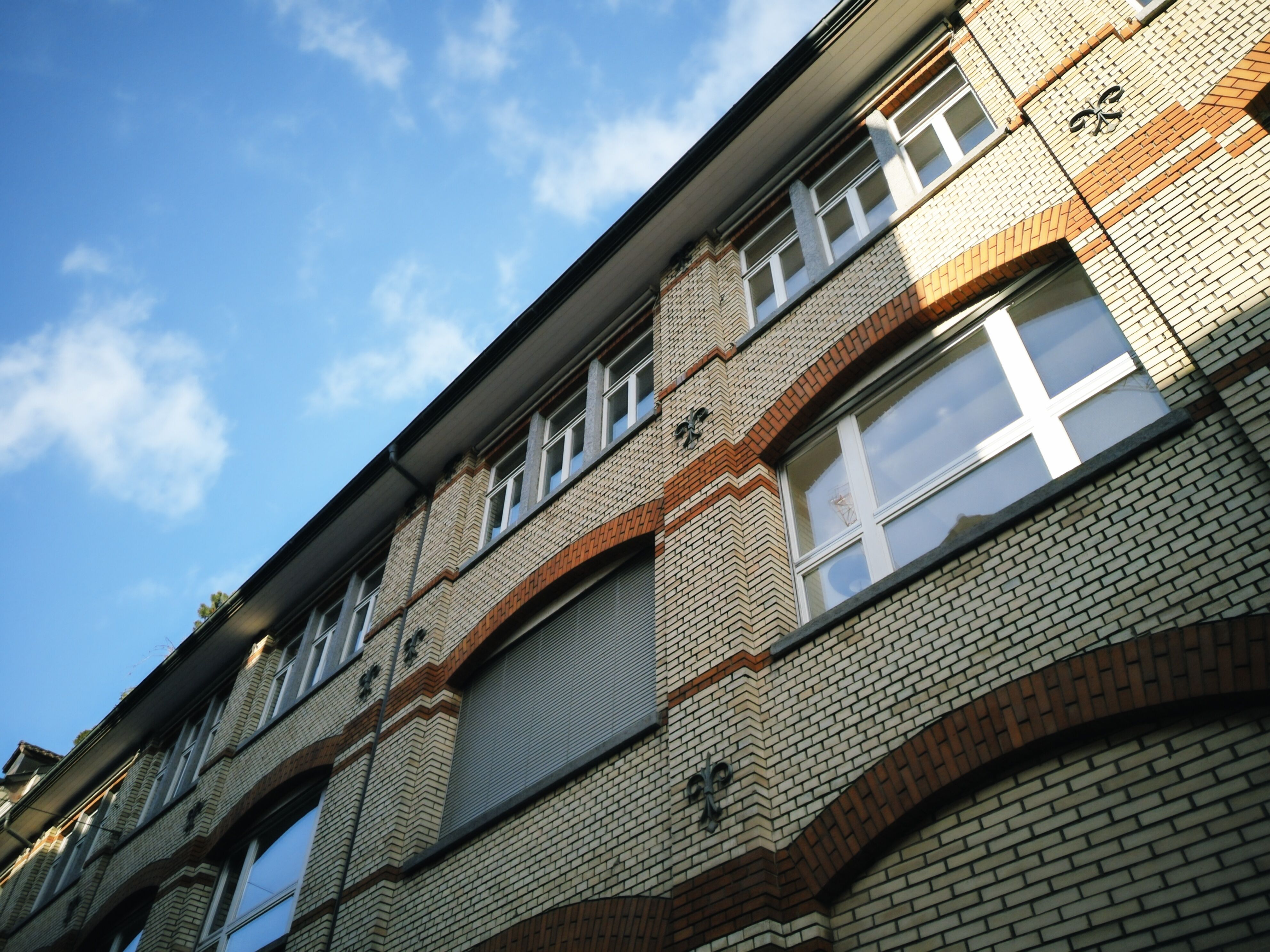}
\end{figure}
\end{landscape}

\begin{landscape}
\begin{figure}
\centering
\includegraphics[height=0.95\textheight]{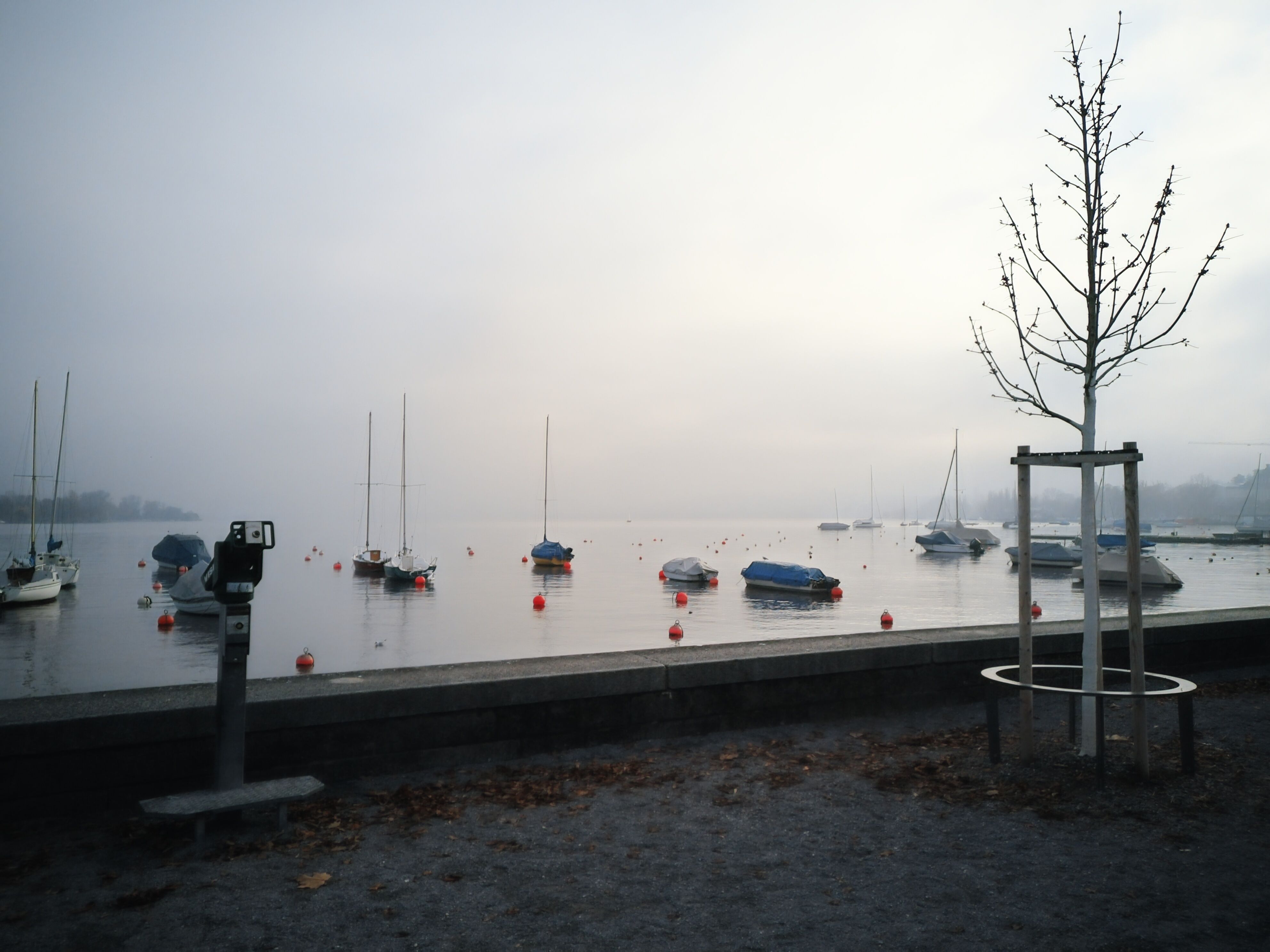}
\end{figure}
\end{landscape}

\end{document}